\pdfoutput=1

\documentclass[11pt]{article}

\usepackage[final]{acl}

\usepackage{times}
\usepackage{latexsym}
\usepackage{booktabs}
\usepackage{listings}
\usepackage{enumitem}
\usepackage{subcaption}
\setlist{nosep}

\usepackage[T1]{fontenc}

\usepackage[utf8]{inputenc}

\usepackage{microtype}

\usepackage{inconsolata}

\usepackage{graphicx}
\usepackage{multirow}
\usepackage{tcolorbox}
\usepackage{fancyvrb}

\usepackage{tcolorbox}
\tcbuselibrary{listings, skins, breakable}

\newcommand{\dataset}{RuCCoD}
\newcommand{\datasettt}{RuCCoD-DP}
\title{\dataset: Towards Automated ICD Coding in Russian}

\author{Aleksandr Nesterov\textsuperscript{\rm 1}\thanks{These authors contributed equally to this work.}, Andrey Sakhovskiy\textsuperscript{\rm 2,3}$^*$, Ivan Sviridov\textsuperscript{\rm 4}$^*$, Airat Valiev\textsuperscript{\rm 5}
\\ {\bf Vladimir Makharev\textsuperscript{\rm 1,6}, Petr Anokhin\textsuperscript{\rm 1}, Galina Zubkova\textsuperscript{\rm 4},
Elena Tutubalina\textsuperscript{\rm 1,2,7}} 
\\
\textsuperscript{\rm 1} AIRI
\textsuperscript{\rm 2} Sber AI
\textsuperscript{\rm 3} Skoltech
\textsuperscript{\rm 4} Sber AI Lab\\
\textsuperscript{\rm 5} HSE University 
\textsuperscript{\rm 6} Innopolis University 
\textsuperscript{\rm 7} ISP RAS Research Center for Trusted AI\\
\\}

\begin{document}
\maketitle
\begin{abstract}
This study investigates the feasibility of automating clinical coding in Russian, a language with limited biomedical resources. We present a new dataset for ICD coding, which includes diagnosis fields from electronic health records (EHRs) annotated with over 10,000 entities and more than 1,500 unique ICD codes. This dataset serves as a benchmark for several state-of-the-art models, including BERT, LLaMA with LoRA, and RAG, with additional experiments examining transfer learning across domains (from PubMed abstracts to medical diagnosis) and terminologies (from UMLS concepts to ICD codes).
We then apply the best-performing model to label an in-house EHR dataset containing patient histories from 2017 to 2021. Our experiments, conducted on a carefully curated test set, demonstrate that training with the automated predicted codes leads to a significant improvement in accuracy compared to manually annotated data from physicians. We believe our findings offer valuable insights into the potential for automating clinical coding in resource-limited languages like Russian, which could enhance clinical efficiency and data accuracy in these contexts. Our code and dataset are available at \url{https://github.com/auto-icd-coding/ruccod}.
\end{abstract}

\section{Introduction}

The explosion of medical data driven by technology and digitalization presents a unique opportunity to enhance healthcare quality. With the adoption and implementation of electronic health records (EHRs), accurate and timely data utilization is crucial for effective treatment and disease management. Central to this process is the assignment of International Classification of Diseases (ICD) codes, which is essential for medical documentation, billing \cite{sonabend2020automated}, insurance  \cite{park2000accuracy}, and research \cite{bai2018interpretable,lu2022context,ijcai2019p0825}.

Although ICD code assignment is crucial for EHRs, it poses significant challenges. Human coders must navigate a wide array of medical terminology, subjective interpretations, and time pressures, all while staying updated with constantly changing classification standards \cite{Burns2012SystematicRO, OMalley2005MeasuringDI, Cheng2009TheRA}.
Coding errors can lead to misdiagnosis, ineffective treatment, diminished trust in the healthcare system, and negative public health outcomes. Furthermore, errors in manual coding in the ICD system, result in financial repercussions, accounting for 6.8\% of the total payments \cite{manchikanti2002implications}.

\begin{figure}[t!]
    \centering
    \includegraphics[width=0.50\textwidth]{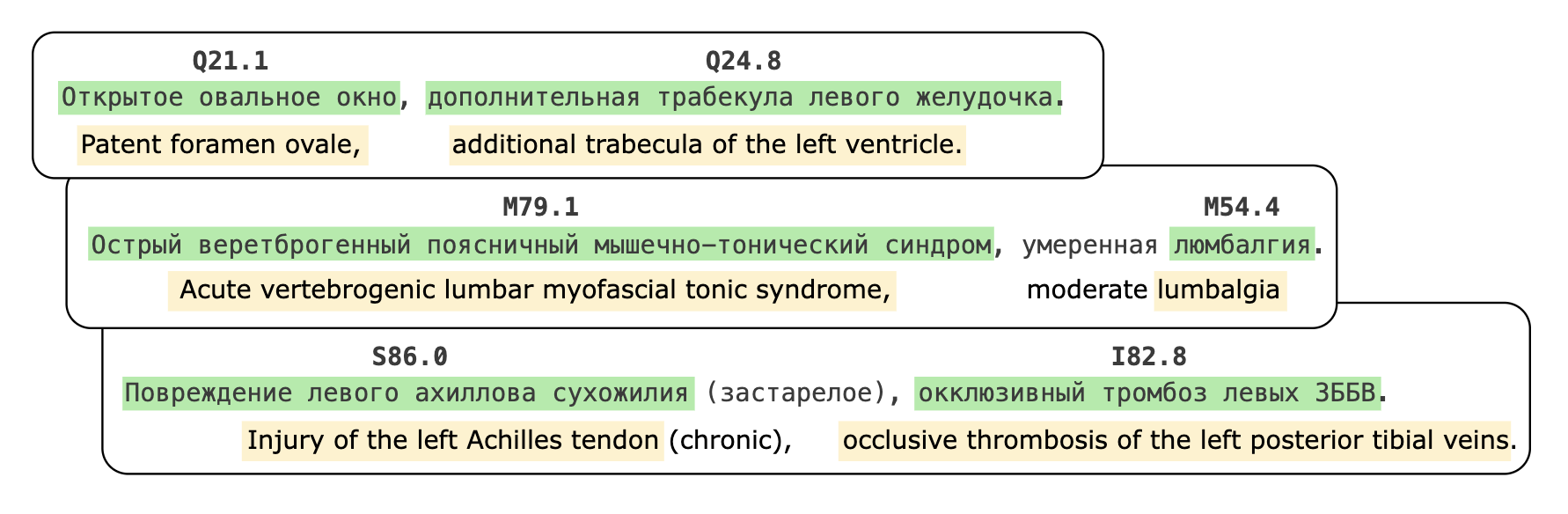}
    \caption{Examples of ICD code assignments by annotators: each entity in green is annotated with its ICD code above and its English translation (in yellow).}
    \label{fig:brat}
    \vspace{-0.2cm}
\end{figure}


\begin{figure*}[t!]
    \centering
    \begin{subfigure}[t]{0.48\textwidth}
        \centering
        \includegraphics[width=\linewidth, height=5.5cm]{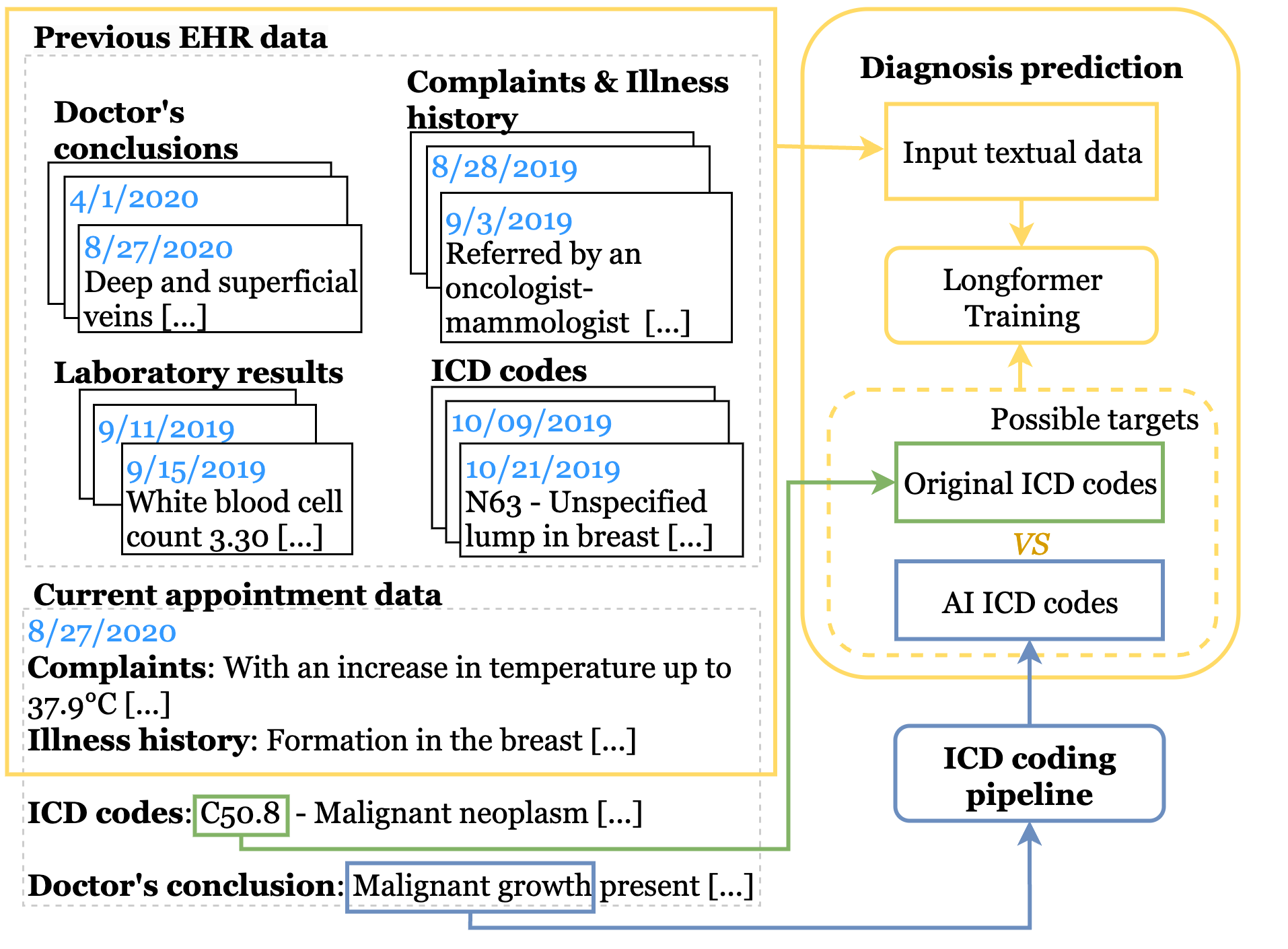}
        \label{fig:schema_a}
    \end{subfigure}
    \hfill
    \begin{subfigure}[t]{0.48\textwidth}
        \centering
        \includegraphics[width=\linewidth, height=5.5cm]{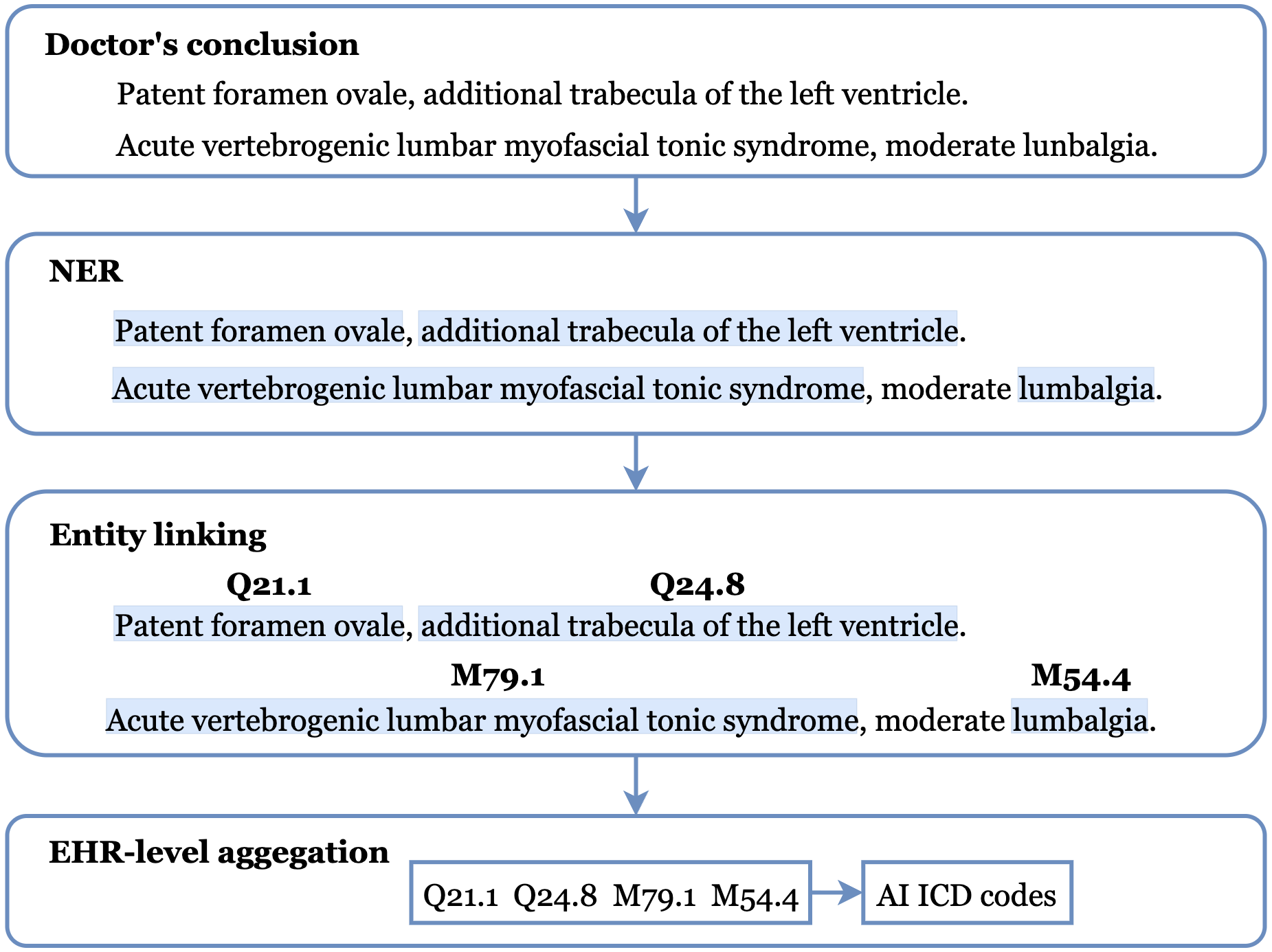}
        \label{fig:schema_b}
    \end{subfigure}
    \caption{An overview of \textit{ICD coding} and \textit{diagnosis prediction} tasks. \textit{Diagnosis prediction} (\textbf{left}, yellow) is to predict diagnoses as ICD codes based on prior EHR data and current visit details, excluding a doctor’s conclusion. Two training label types are compared: (i) \textit{original} ICD codes (manually assigned by physicians, shown in green) and (ii) \textit{AI ICD codes} (automatically generated by our \textit{ICD coding} pipeline, shown in blue). \textit{ICD coding} pipeline (\textbf{right}) extracts and links diseases from a doctor-written conclusion, then assigns the deduplicated codes to an EHR. }
    \label{fig:schema}
    \vspace{-0.1cm}
\end{figure*}

Schematic description of the \textit{ICD coding} and \textit{diagnosis prediction} tasks (left panel, blue and yellow, respectively), along with a detailed illustration of the \textit{ICD coding} pipeline (right panel).


Despite extensive research on ICD coding using neural networks \cite{li2020icd,zhou2021automatic,yuan2022code,baksi2024medcodergenerativeaiassistant, boyle2023automatedclinicalcodingusing,mullenbach2018explainable,cao2020hypercore,yuan2022code,yang2022knowledge,huang-etal-2022-plm}, significant challenges persist for non-English languages. These include low inter-coder agreement, limited labeled data, variability in clinical notes, the hierarchy of ICD codes, and reliance on incomplete input data. In addition, clinical environments in low-resource languages exhibit a shortage of certified medical coders, forcing physicians to assign ICD codes themselves, leading to inconsistencies and low-quality coding. To address these issues, we introduce a novel dataset for automatic ICD coding in Russian and explore automated approaches to assist physicians in the coding process.

Previous studies have primarily focused on English-language datasets, specifically MIMIC-III/IV \cite{goldbergerPhysioBankPhysioToolkitPhysioNet2000, johnsonMIMICIVFreelyAccessible2023}. 
Despite being one of the top ten languages in terms of concept name count within the Unified Medical Language System (UMLS) \cite{bodenreider2004unified}  biomedical metathesaurus, Russian remains underdeveloped in the clinical domain. The Russian segment of the UMLS comprises only 1.96\% of the vocabulary and 1.62\% of the source counts found in the English UMLS \cite{nihumls}.
Recent corpora, such as RuCCoN \cite{nesterov-etal-2022-ruccon} and NEREL-BIO \cite{NERELBIO}, focus on concepts within the Russian UMLS.

In this work, we explore two closely related tasks: \textbf{ICD coding} and \textbf{Diagnosis Prediction} (\textbf{DP}). Both tasks are designed to assist physicians by standardizing and automating the diagnosis process into ICD codes, especially in low-resource clinical settings where professional coders are not available. As seen in Fig.~\ref{fig:schema}, the tasks take non-overlapping input and complement each other: \textit{ICD coding} normalizes a free-form doctor's diagnosis conclusion into a set of relevant ICD codes while the DP task is to directly predict ICD-agreed diagnoses from EHRs in one pass without relying on the doctor's textual diagnosis conclusion. Although  we formulate \textit{ICD coding} as an entity normalization task and \textit{DP} as multilabel classification, both tasks are sometimes referred to as \textbf{ICD coding}.
Unlike prior classification-based \textit{ICD coding} research~\citep{li2020icd,vu2020label,wang2024icdxml}, we explore a more challenging scenario in which a diagnostic model, acting as an independent medical expert, predicts diagnoses from patient data only without relying on the doctor's diagnosis conclusion.
Thus, we term the classification task \textbf{diagnosis prediction} as it better reflect the problem's nature and does not create a confusion with linking-based \textit{ICD coding}~\cite{lavergne2016dataset, neveol2017clef, coutinho2022transformer}. 

For \textit{ICD coding}, we present \textbf{\dataset{}} (\textbf{Ru}ssian I\textbf{C}D \textbf{Co}ding \textbf{D}ataset), a novel dataset in Russian, labeled by medical professionals based on concepts from the ICD-10 CM (Clinical Modification) system (Sec. \ref{sec:ruccod_icd_coding}). Second, we establish a \textbf{comprehensive benchmark} for state-of-the-art models, including a BERT-based \cite{devlin-etal-2019-bert} pipeline for information extraction, a LLaMa-based \cite{touvron2023llama} model with Parameter Efficient Fine-Tuning (PEFT) and with retrieval-augmented generation (RAG).
Furthermore, we evaluate transfer learning of models trained on UMLS concepts and similar biomedical datasets (PubMed abstracts \cite{NERELBIO}, clinical notes \cite{nesterov-etal-2022-ruccon}. The results suggest that the ICD's fine-grained hierarchical structure hinders generalization from other clinical sources (Sec. \ref{sec:expcoding}).

For \textit{diagnosis prediction}, we perform a set of experiments on \textbf{\datasettt{}}, a large in-house dataset of 865k EHRs from 164k patients.
When training a diagnostic model, we experiment with ICD codes assigned by doctors during patient appointments, as well as the \textbf{AI-assigned ICD codes} (Sec. \ref{sec:expdiag}), that is, diagnoses assigned by automatically linking an EHR diagnosis conclusion with a top-performing \textit{ICD coding} model on RuCCoD (see Fig.~\ref{fig:schema}).
Our experiments have revealed that pre-training on automatically assigned ICD codes gives a huge weighted F1-score growth of 28\% for \textbf{diagnosis prediction} compared to physician-assigned ICD codes indicating the difficulty of ICD-guided diagnosis formalization for physicians and great potential of AI-aided diagnosing.
Our work provides a foundation and guidance for ICD-related research in low-resource clinical languages.

\section{ICD-Related Tasks}\label{sec:icd_tasks}

\paragraph{Task: ICD coding} is akin to Entity Linking (EL), where the objective is to assign a set of unique ICD codes to the latest patient appointment  based on textual diagnosis conclusion written by a doctor. The task aims to help a physician normalize diagnosed diseases to a set of codes from the complex formal ICD hierarchy.
We model the \textit{ICD Coding} as an \textit{information extraction} pipeline with three components: (1) \textit{Nested Entity Recognition} (NER) and (2) EL followed by (3) \textit{EHR-level code aggregation}. 
Step (3) minimizes NER influence on pipeline metrics by omitting NER spans.
The approach aligns with real-world ICD applications, where the primary objective is accurate assignment of ICD codes (i.e., disease recognition), and imprecise NER outputs are not impactful.




\paragraph{ICD Coding: EHR-level Code Aggregation} Given an EHR, we perform EL on NER predictions. Let $L_p = (c^p_1, c^p_2,\dots,c^p_n)$ and $L_t = (c^t_1, c^t_2,\dots,c^t_m)$ denote the lists of predicted and ground truth ICD codes, respectively. After a standard NER+EL pipeline, each list may contain multiple mentions of the same disease (i.e., $c^t_i = c^t_j$ for $i \neq j$). The presence of duplicate disease mentions prevents entity-level metric aggregation, as they are erroneously counted  multiple times, thereby introducing bias into the performance evaluation. We remove duplicates from both lists, resulting in unique code sets $S_p$ and $S_t$ such that $ c^p_i \neq c^p_j$ and $c^t_i \neq c^t_j, \forall i \neq j$.
Finally, micro-averaged classification metrics are calculated from True Positives ($TP$), False Positives ($FP$) and False Negatives ($FN$): $TP=S_p \cap S_t$; $FP=S_p \setminus S_t$; $FN=S_t \setminus S_p$.




\paragraph{Diagnosis Prediction}
is a multi-label classification task that outputs likely diagnoses (ICD codes) for the current doctor appointment from a patient's \textit{past medical history}, including complaints, test and examination results from previous appointments.
In our study, each EHR contains a doctor's diagnosis conclusion. A major challenge for ICD-grounded applications is that this conclusion is a free-form text, and its normalization to ICD might introduce sensitive errors. Conversely, automatic \textit{Diagnosis Prediction} is constrained to output ICD-compliant diagnoses by task design.

\paragraph{ICD Coding vs. Diagnosis Prediction} While \textit{ICD Coding} only observes the current appointment's diagnosis conclusion written by a doctor, the goal of \textit{Diagnosis Prediction} is to actually write the diagnosis conclusion (i.e., make an \textbf{AI diagnosis conclusion}). Here, the motivation is to offer a doctor an independent, AI-driven opinion, potentially beneficial for decision-making in complex cases. Hence, the \textbf{two tasks are complementary by design}, using non-overlaping EHR parts: \textit{ICD Coding} leverages the latest diagnosis while \textit{Diagnosis Prediction} observes an entire patient's history except for the latest diagnosis conclusion.






\section{ICD Datasets}\label{sec:all_icd_datasets}

\subsection{RuCCoD: ICD Coding Dataset}\label{sec:ruccod_icd_coding}


For \textbf{ICD coding}, we release \textbf{\dataset}, the first dataset of Russian EHRs with disease entities manually linked to ICD-10.
In this section, we describe the data collection and annotation pipeline and provide important statistics.

\begin{table}[t!]
\centering
\begin{tabular}{lll}
\hline
                                    & \textbf{Train} & \textbf{Test} \\ \hline
\# of records                   & 3000               & 500               \\
\# of assigned entities         & 8769               & 1557              \\
\# of unique ICD codes          & 1455               & 548               \\
Avg. \# of codes per record & 3                  & 3                 \\ \hline
\end{tabular}
\caption{Statistics for the \dataset{} training and testing sets on ICD coding of diagnosis.}
\label{tab:ruccod_stats}
\vspace{-0.2cm}
\end{table}

\begin{figure}[t!]
    \centering
    \includegraphics[width=0.47\textwidth]{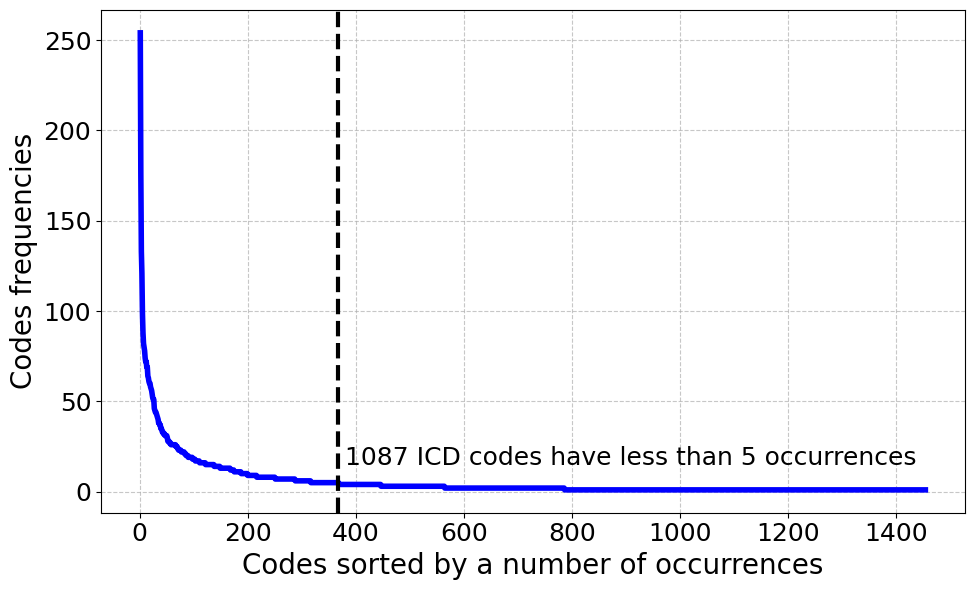}
    \caption{Distribution of ICD code frequencies in the \dataset{} train set.}
    \label{fig:codes_stats}
\end{figure}

\begin{table*}[t!]
\centering
    \scalebox{0.87}{\begin{tabular}{@{}lccccc@{}}
    \toprule
  & \textbf{Original Dataset} & \textbf{Linked Dataset} & \textbf{Common Test Set } \\ \midrule
Number of records                    & 865539           & 865539         & 494                   \\
Number of unique patients            & 164527           & 164527         & 450                   \\
Number of unique ICD codes           & 3546             & 3546           & 394                   \\
Avg. number of ICD codes per patient & $3\pm2$                & $5\pm2$              & $4\pm2$                     \\ 
    Avg. number of EHR records before current appointment & (15, 36, 73) & (15, 36, 73) & (17, 36, 77)                     \\ 
    Avg. length of EHR records per one appointment & (77, 167, 316) & (77, 167, 316)              & (86, 176, 320)                     \\ 
    Patient's age & (59, 67, 74) & (59, 67, 74)              & (60, 67, 75)                     \\ 
    Percentage of male patients & 69 & 69              & 71                     \\ 
    \bottomrule
    \end{tabular}} 
    \caption{Statistics for the randomly split training and testing sets of \datasettt{} for diagnosis prediction. Values in brackets show the 25th, 50th, and 75th percentiles.}
    \label{tab:datastats}
\end{table*}

\paragraph{Data Collection}

As a \textbf{source} for \dataset, we utilize diagnosis conclusions from the records of a major European city's Medical Information System. 
Before starting the annotation process, we implemented a meticulous de-identification protocol to protect data privacy. Medical professionals invited to annotate the dataset first conducted a comprehensive manual review of all diagnoses. Their task was to identify and remove any personal or identifiable information manually. This thorough process guarantees compliance with privacy regulations and ensures the dataset is suitable for research use.

\paragraph{Annotation Process and Principles}
The labeling team consisted of three highly qualified experts with advanced education in different fields of medicine, two of whom hold Ph.D. degrees, with every annotation further validated by a fourth expert, a Ph.D. holder in medicine. Grounded in the ICD-10 CM (Clinical Modification) system, the team aimed to identify all nosological units in a diagnosis conclusion and assign the most accurate ICD code to each. An annotation example is shown in Fig.~\ref{fig:brat}.
The dataset was randomly split into 3,000 training and 500 testing records. Each expert independently annotated 1,000 training records for diverse labeling, while all three annotated the same 500 test records for consistency. An ICD code was accepted if at least two annotators agreed. Annotation guidelines and details on inter-annotator agreement are given in Appx. \ref{app:ann_gui}.

\paragraph{Dataset Statistics}

Statistics of train and test splits of the RuCCoD dataset are provided in Tab.~\ref{tab:ruccod_stats}. Despite the large number of ICD codes, especially in the training set, their distribution is uneven. Fig.~\ref{fig:codes_stats} shows the distribution of ICD codes within the RuCCoD train set. While a small number of codes dominate the dataset, appearing from 50 to 250 occurrences, most codes are rare, with 1,087 codes occurring fewer than 5 times. This stark disparity underscores the challenges of dealing with real-world medical data, where frequent diagnoses are well-represented, but rare conditions remain significantly under-sampled.




\subsection{RuCCoD-DP: Diagnosis Prediction Dataset}\label{sec:ruccod-dp}

To explore AI-guided \textbf{Diagnosis Prediction}, we collect \textbf{\datasettt{}} (\textbf{\dataset{}} for \textbf{D}iagnosis \textbf{P}rediction), a corpus of real-world EHRs.

\paragraph{Dataset Construction}

\datasettt{} includes doctor appointments from 2017 to 2021, divided into four parts: (i) patient complaints and anamnesis, (ii) lab test results, (iii) appointment summary (including assigned ICD codes), and (iv) appointment history. Although \dataset{} and \datasettt{} share a common source, we ensured that both sets contain no overlapping patients and, consequently, no overlapping appointments. 

\paragraph{Paired Human-AI ICD Codes} ICD has a fine-grained disease hierarchy introducing a significant challenge even for a qualified doctor to formalize a correctly diagnosed disease . For instance, a \textit{H10 Conjunctivitis} disease group has 8 specifications including: \textit{H10.0 mucopurulent}, \textit{H10.1 acute atopic}, \textit{H10.2 other acute}, and \textit{H10.3 unspecified acute} conjunctivitis. 
Thus, doctor-assigned ICD codes in real-world EHRs can expose substantial errors even if a general disease is diagnosed correctly.
To address the issue, we consider two ICD code sets for each EHR: (i) real-world ICD codes originally written by physicians within the EHR (\textbf{doctor-assigned codes}); (ii) automatically assigned ICD codes predicted by a neural model trained on \textit{\dataset{}} (\textbf{AI codes}). 
\textbf{AI codes} (i.e., \textit{AI-assigned diseases}) are assigned to an EHR by applying our top-performing BERT-based NER+EL \textit{ICD Linking} pipeline (Tab.~\ref{tab:transfer}) to the EHR's doctor's real-world diagnosis conclusion (see Fig.~\ref{fig:schema}). Our pipeline extracts diseases (NER) and links (EL) them to ICD codes and then the found diseases are assigned to the given EHR labels for \textit{ICD Coding}. Thus, \textit{AI codes} are designed to aid in the formalization of the human-written diagnosis to the ICD code set while relying only on the written conclusion of the physician. Notably, the two coding types rely on the same underlying free-form diagnosis conclusions.

\paragraph{Original and Linked \datasettt{}} We will refer to \datasettt{} variations sharing the same appointments yet different in ICD code assignment method (either \textit{doctor-assigned} or \textit{AI-based}) as \textbf{original} and \textbf{linked} datasets, respectively. In other words, a single textual appointment entry has two distinct labels sets. To prevent ICD codes distribution shift between \textit{original} and \textit{linked} data, we retained the ICD codes overlapping between these two sets. For each appointment sample, its textual input included the concatenation of chronologically sorted all prior appointments. 

\paragraph{Diagnosis Prediction Test Set}
The collection of two sets of labels allows exploration of whether manual or generated ICD labels are more reliable for model training. For a fair comparison of the labeling approaches, we developed a \textit{common test set} from a subset of the original appointment dataset's test set. We formed it by selecting a subset from the test part of the original appointment dataset. For annotation, we adopted the same annotation methodology as for the \dataset{} dataset (Sec.~\ref{sec:ruccod_icd_coding}). 
The mean pairwise JSC across annotators was 0.331. The final statistics for \textit{original}, \textit{linked} datasets as well as for the \textit{common test set} is summarized in Tab.~\ref{tab:datastats}.

\begin{table*}[t!]
\centering
\begin{tabular}{lrrrr}
\toprule
\multirow{1}{*}{\textbf{Model}} & \multicolumn{1}{c}{\textbf{Precision}}  & \multicolumn{1}{c}{\textbf{Recall}} & \multicolumn{1}{c}{\textbf{F-score}} & \multicolumn{1}{c}{\textbf{Accuracy}}  \\
\midrule 
\multicolumn{5}{c}{\textbf{Supervised with various corpora for NER and EL}} \\
\midrule
BERT, NER: NEREL-BIO + \dataset, EL: \dataset & \textbf{0.512} & 0.529 & 	0.520 &  0.352 \\
BERT, NER: RuCCoN + \dataset, EL: \dataset & 0.471 & \textbf{0.543} & 0.504 &  0.337 \\
BERT, NER: \dataset, EL: \dataset & 0.510 & 0.542 & 	\textbf{0.525} &  \textbf{0.356} \\\midrule

\multicolumn{5}{c}{\textbf{LLM with RAG (zero-shot with dictionaries)}} \\
\midrule
LLaMA3-8b-Instruct, NEREL-BIO & 0.059 & 0.053 & 0.056 & 0.029 \\ 
LLaMA3-8b-Instruct, RuCCoN & 0.164 & 0.15 & 0.157 & 0.085 \\
LLaMA3-8b-Instruct, ICD dict. & 0.379 & 0.363 & 0.371 & 0.228 \\
LLaMA3-8b-Instruct, ICD dict. + \dataset & 0.465 & 0.451 & 0.458 & 0.297 \\
\midrule
\multicolumn{5}{c}{\textbf{LLM with tuning}} \\
\midrule
Phi3\_5\_mini, ICD dict. & 0.394 & 0.39 & 0.392 & 0.244 \\
Phi3\_5\_mini, ICD dict. + \dataset{} & 0.483 & 0.477 & 0.48 & 0.316 \\
Phi3\_5\_mini, ICD dict. + BERGAMOT & 0.454 & 0.448 & 0.451 & 0.291 \\
\bottomrule
\end{tabular}\caption{Entity-level code assignment metrics on \dataset's test set. The best results are highlighted in \textbf{bold}. We also refer to Appx. \ref{appx:cross_terminology_transfer}, \ref{app:llm_rag_res}, \ref{app:llm_tuning_res} on more experiments with different LMs, corpora, and terminologies.}\label{tab:transfer}
\end{table*}

\section{ICD Coding Evaluation}\label{sec:expcoding}

For \textbf{ICD coding} experiments, we experiment with the following approaches: 1) a fine-tuned BERT-based pipeline for information extraction, 2) a large language model (LLM) with Parameter-Efficient Fine-Tuning (PEFT), and 3) LLM with retrieval-augmented generation (RAG). All three systems use the same dictionary, with 17,762 pairs of codes and diagnoses (refered to as \textit{ICD dict}) compiled from the Ministry of Health data. In addition, LLM-based systems used a train set as a dictionary as well. See the Appx. \ref{appx:impl_details} for a list of the LLMs used. See related work in Appx. \ref{appx:rel}.

\subsection{Models}


\paragraph{BERT-based IE Pipeline} Our Information Extraction (IE) pipeline uses sequential NER and EL modules. The NER module, employing a softmax layer, extracts relevant entities, and the EL module then links these entities to ICD codes based on semantic similarity with ICD dictionary entries.
For NER, we utilize the pre-trained RuBioBERT~\cite{yalunin2022rubioroberta}, 
and for EL, we employ the multilingual state-of-the-art models SapBERT~\citep{liu-etal-2021-self,liu-etal-2021-learning-domain}, CODER~\citep{CODER-BERT}, and BERGAMOT~\cite{sakhovskiy-etal-2024-biomedical}. 
In EL, entities and vocabular disease names are encoded by a BERT-based model with average pooling, and the top-$k$ concepts are selected for each entity based on the nearest Euclidean distance.
We fine-tuned models on EL train sets via synonym marginalization proposed in \emph{BioSyn} \cite{sung-etal-2020-biomedical}. For more details, see Appx~\ref{appx:impl_details}.

\paragraph{LLMs with PEFT}  
We explored the capabilities of LLMs for clinical coding using PEFT with Low-Rank Adaptation (LoRA) \cite{hu2021lora}. The pipeline included two steps: NER and EL, following the structure of BERT-based IE pipeline described earlier. 
For NER stage, models were fine-tuned on \dataset{} using task-specific prompts (Appx.~\ref{ner-prompt}).
The predictions were validated by exact string matching and Levenshtein distance with a threshold $\leq2$ chosen empirically to optimize the robustness of the spelling without overcorrecting semantically distinct entities.
For EL, a RAG approach was implemented to link extracted entities to ICD codes. The retrieval component was built using three strategies: (1) BGE embeddings \cite{chen2024bgem3embeddingmultilingualmultifunctionality} on the \textit{ICD dict}, (2) BGE embeddings on the \textit{ICD dict} combined with \dataset{} training entities, and (3) BERGAMOT embeddings \cite{sakhovskiy-etal-2024-biomedical} fine-tuned on \dataset{} with the \textit{ICD dict}.

We adopted the FAISS index \cite{douze2024faiss} to retrieve the top-15 most similar dictionary entries for each entity extracted in the NER stage.
The final ICD code was assigned using an LLM to select the closest match from the retrieved candidates (prompt in Appx.~\ref{sec:prompt}). To address class imbalance, diagnosis lists were shuffled during training, forcing models to learn contextual code-discrimination. Fine-tuning parameters followed standard LoRA configurations (Tab.~\ref{tab:hp}, Appx.~\ref{appx:impl_details}).

\paragraph{Zero-shot LLM with RAG}  
As an ablation study, we evaluated the same pipeline as in the PEFT stage but without fine-tuning to isolate the LLMs' inherent capabilities. We used only the fine-tuned BERGAMOT embeddings from strategy (3) for retrieval, retaining the FAISS index and prompts (Appx.~\ref{sec:prompt}). The LLM selected ICD codes from retrieved candidates if no direct match was found, replicating the EL process from the PEFT stage. This setup allowed us to quantify the contribution of fine-tuning versus zero-shot inference.

\subsection{Evaluation Methodology}


On \dataset, our evaluation includes conventional NER and EL as well as end-to-end document-level code assignment with EHR-level code aggregation (Sec.~\ref{sec:icd_tasks}).
To recall, document-level metrics is an entity position-agnostic NER+EL task composition with explicitly removed EHR-level ICD code duplicates.
For instance, a language model successfully diagnoses a patient by assigning the correct ICD code when it finds at least one of three mentions of the corresponding ICD disease within an EHR.
For all three tasks, we report accuracy and the micro-averaged precision, recall, F1-score. 

Following prior research~\citep{phan-etal-2019-robust,wright2019normco,liu-etal-2021-self,CODER-BERT,sakhovskiy-etal-2024-biomedical}, we use a retrieval-based EL approach and evaluate retrieval accuracy:
$acc@k = 1$ if a correct ICD code is retrieved at rank $\leq k$. We consider two evaluation scenarios: (i) \textit{strict} score assessing exact match between a predicted ground truth codes; (ii) \textit{relaxed} score with each code being truncated to higher-level disease group (e.g., \textit{H10.0 Mucopurulent conjunctivitis} is truncated to \textit{H10 Conjunctivitis}).

\subsection{Results}


\subsubsection{Transfer Learning}
First, we performed cross-domain experiments on EL to see how variability in entities and terminology affects the performance. Since UMLS includes the ICD system, we automatically map UMLS CUIs to ICD codes for evaluation. Cross-domain transfer results with entity linking models on \dataset, RuCCoN, NEREL-BIO and their union are presented in detail in Appx.~\ref{appx:cross_terminology_transfer}. The evaluation has revealed the following key observations. 

\paragraph{Maleficent Cross-Domain Vocabulary Extension} While extension of ICD vocabulary consistently gives a slightly improved acc@1 in a zero-shot setting, additional synonyms introduce severe noise in a supervised setting. Specifically, a significant drop of 8.1\%, 8.4\%, and 14.3\% acc@1 is observed for SapBERT, CODER, and BERGAMOT, respectively. Even in an unsupervised setting, vocabulary extension drops acc@5 by 5.2\% and 6.8\% for SapBERT and BERGAMOT, respectively. The results highlight the specifity of ICD terminology within the medical domain.

\paragraph{Limitations of Cross-Terminology Transfer} Both training on RuCCoN and NEREL-BIO as well as merge of these corpora with \dataset{}  do not lead to improvement over zero-shot coding (Tab.~\ref{tab:res_entity_linking_transfer}, Appx.~\ref{appx:cross_terminology_transfer}). Thus, training on other EL corpora normalized to the UMLS vocabulary does not transfer to our \dataset{} dataset normalized to ICD vocabulary. The finding indicates the specificity and high complexity of hierarchical ICD coding within the EL task.


\paragraph{Complexity of Fine-Grained ICD Coding} The 15\% gap in acc@1 between the \textit{strict} and \textit{relaxed} evaluation (Tab.~\ref{tab:res_entity_linking_transfer}, Appx.~\ref{appx:cross_terminology_transfer}) shows the challenging nature of  semantically similar diseases within the same therapeutic group. The finding further illustrates the underlying challenge for human ICD coders: correct disease specification with an appropriate code is a far more complex task than general disease identification.


\paragraph{Transfer learning for NER is Feasible}
A NER model trained on the disease-related entities from NEREL-BIO gained an F1 score of 0.62 on \dataset{}'s test set.
The model trained on a combined dataset of NEREL-BIO and \dataset{} achieved scores of 0.72. Similar results were observed with RuCCoN. We also evaluated BINDER, which uses a RuBioBERT backbone and treats NER as a representation learning problem by maximizing similarity between vector representations \cite{zhangoptimizing}.
However, BINDER's performance was 1.5\% lower than RuBioBERT's, which gained the best F1 score of 0.77 with a softmax classifier.
NER transfer for disease entities is significantly better than for entity linking (EL), with the best results obtained from \dataset{} (full results are in Appx. \ref{appx:ner-bert}).

\subsubsection{End-to-end ICD coding}
In the next experiments, we evaluated an end-to-end ICD coding quality on raw texts, in which models were fine-tuned on either RuCCoN or NEREL-BIO or utilized entity dictionaries from these datasets, are presented in Tab.~\ref{tab:transfer}. As seen from the results, training on datasets from other domains gives limited performance and the best ICD coding results are observed for models trained with ICD data from RuCCoD data on all three set-ups.

Extended RAG results are in Appx. \ref{app:llm_rag_res}. Fine-tuning LLMs improves performance across all tasks, exceeding LLM + RAG results in zero-shot settings. Use of \dataset{} significantly enhances metrics compared to approaches that rely solely on the ICD dictionary or embeddings. Llama3-Med42-8B and Phi3\_5\_mini are the most effective models after PEFT tuning (see Appx. \ref{app:llm_tuning_res}).

\begin{figure}[t!]
    \centering
    \includegraphics[width=0.45\textwidth]{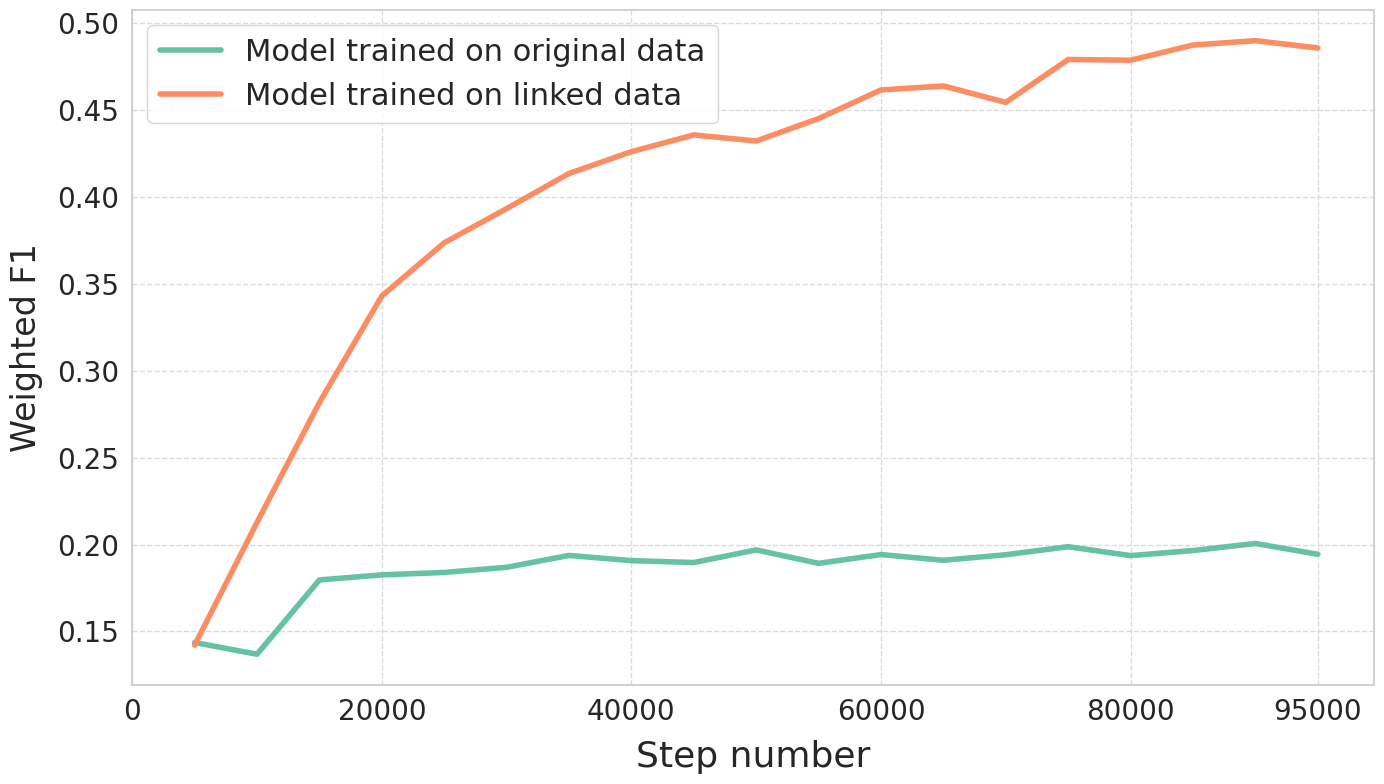} 
    \caption{Comparison of weighted F1 scores on the common test set for models trained on \textit{original} and \textit{linked} datasets at different training steps.}
    \label{fig:macro_f1} 
\end{figure}

\section{Diagnosis Prediction Evaluation}\label{sec:expdiag}

For the \textit{Diagnosis Prediction} task, we explore whether automatic relabeling of ICD codes assigned to an EHR increases the robustness and reliance of a diagnostic model. To address this, we train two diagnostic models on \datasettt{}'s (i) \textit{original} and (ii) \textit{linked} independently. These two models learn from the same EHRs but use different target ICD labels (\textit{doctor-assigned} and \textit{AI-based} codes, respectively). Given that both annotation types are produced from human-written diagnosis conclusions, our experimental design compares two methodologies for formalizing a verbose diagnostic conclusion into a set of ICD codes.

\subsection{Experimental Set-up}


\paragraph{Model} We chose the Longformer architecture~\cite{DBLP:journals/corr/abs-2004-05150} due to its strong performance in clinical tasks~\cite{Edin2023AutomatedMC}. Our Longformer model is initialized from a BERT model pre-trained using private EHRs from multiple clinics and further pre-trained on extended sequences. Training details are in Appx.~\ref{appx:impl_details}.

\paragraph{Evaluation}

In our experiments, we evaluate the quality of the models trained on the \textit{original} and \textit{linked} datasets on the \textit{common test set}. Because training datasets contain a much larger and more diverse set of ICD codes than the test set, all evaluation metrics are computed only over the codes in the test set to ensure a consistent and comparable label space.

To address the imbalanced long-tail ICD code distribution in the \textbf{Diagnosis Prediction} task, we adopt the weighted F1 score for overall evaluation, a standard choice in prior work on automated ICD coding~\citep{Johnson2019,DBLP:journals/access/BlancoPC20}. The weight of each class is calculated as the proportion of EHRs sharing the given ICD code in the union of both training datasets. Per-class F1 scores were also measured to explore performance variations across frequent and rare ICD codes.

Beyond the weighted F1, we additionally report micro-averaged confusion components—True Positives (TP), False Positives (FP), False Negatives (FN), and True Negatives (TN). In the multi-label setting, each (record, code) pair is treated as a binary outcome, and these components provide a more granular view of diagnostic sensitivity and specificity than F1 alone.

\subsection{Results}

\begin{figure}[t!]
    \centering
    \includegraphics[width=0.5\textwidth]{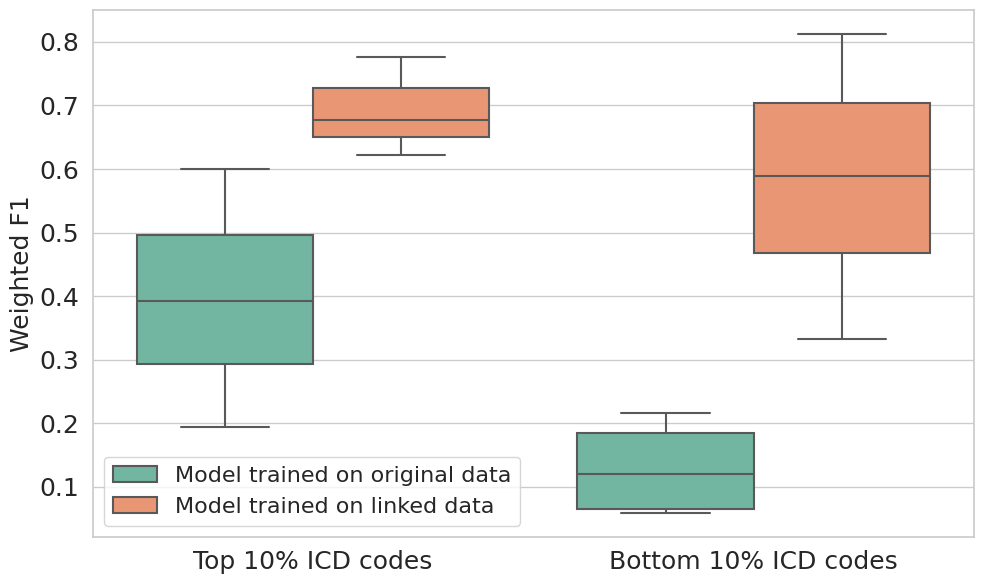} 
    \caption{F1 score distribution for top and bottom 10\% frequent ICD codes in the common test set.}
    \label{fig:top_and_bot} 
    \vspace{-0.25cm}
\end{figure}

\subsubsection{Diagnosis Prediction Learning}
To predict ICD codes from doctor's appointments, we fine-tuned two Longformer models, one using the \textit{original} dataset and the other using the \textit{linked} dataset.
The weighted F1-scores for the two models against the training count are shown in Fig.~\ref{fig:macro_f1}.



\paragraph{AI-based ICD Coding Improves Diagnosing}
As seen from Fig.~\ref{fig:macro_f1}, the model trained on \textit{AI codes} coding (\textit{linked} data) significantly outperforms the one trained on \textit{manual codes} (\textit{original} data) with the peak weighted F1-score of 0.48. The latter quickly reaches its F1-score plateau at 0.2. The huge performance gap of 0.28 highlights the effectiveness of automatic data annotation for model training. The finding reveals the complexity of the ICD-agreed diagnosis prediction task for professional physicians, indicating the need for AI-driven assistance. For instance, an \textit{AI diagnosis conclusion} can serve as an independent second opinion, while the final diagnosis should be determined by a qualified human expert.

\subsection{Diagnosing Stability to Disease Frequency}

Next, we study the diagnosis prediction model's ability to generalize to both frequent and rare disease when trained on \textit{original} and \textit{linked} datasets.

\paragraph{Frequency-Based ICD Test Set Split} The test dataset was split into two parts: the 10\% most frequent ICD codes and the 10\% least frequent ICD codes, with a minimum frequency threshold of 15 instances in the \textit{common} test set for the less frequent group. The stratification approach is designed to align with the distribution of real-world diagnoses assigned and carefully verified by clinicians.



\paragraph{Diagnosing Improvement is Frequency-Robust} Fig.~\ref{fig:top_and_bot} presents the F1 scores spread for individual ICD codes (diseases) grouped by frequency groups. 
The model trained on \textit{linked} data outperforms the one trained on \textit{original} data for both rare and frequent codes.
The  $\sim$6x median F1 score improvement for the bottom 10\% codes (0.6 vs. 0.1) underscores the difficulty of manually assigning ICD codes for infrequent diseases. For frequent codes, the training on \textit{linked} data gives about 0.3 median F1 growth over \textit{original} data ($\sim$0.7 vs 0.4) with a significantly lower score deviation (indicated by smaller interquartile distance). Thus, pretraining on automatically labeled data enhances diagnosis prediction for both rare and common diseases, reducing variability for the latter.


\begin{table}[t!]
\centering
\scalebox{0.8}{
\begin{tabular}{lcccc|cccc}
\toprule
& \multicolumn{4}{c|}{\textbf{Top 10\% codes}} &
  \multicolumn{4}{c}{\textbf{Bottom 10\% codes}} \\
\textbf{Dataset} & FN & FP & TN & TP & FN & FP & TN & TP \\
\midrule
Original & 194 & 146 & 967 & 175 & 36  & 630 & 1284 & 26 \\
Linked   & 118 & 136 & 977 & 251 & 17  & 57  & 1857 & 45 \\
\bottomrule
\end{tabular}}
\caption{Micro-averaged confusion components on the common test set
for the 10\% most frequent (Top) and 10\% least frequent (Bottom) ICD codes, depending on whether the model is trained on \textit{original} or \textit{linked} data.}
\label{tab:confusion_freq}
\vspace{-0.2cm}
\end{table}

We further decompose the results into confusion components to clarify how code re-linking affects different frequency groups. As seen from Table~\ref{tab:confusion_freq}, training on \textit{linked} data reduces False Negatives by 40\% and improves True Positives by 43\% for the most frequent codes compared with the training on the \textit{original} data. For the least frequent codes, we observe a 53\% decrease in False Negatives and a 91\% decrease in False Positives, while True Negatives and True Positives increase by 45\% and 73\%, respectively. This analysis confirms that the improvements from linked training align with the F1 gains, reflecting fewer missed and spurious predictions and more correct detections across both frequent and rare codes.



\begin{figure}[t!]
    \centering
    \includegraphics[width=0.5\textwidth]{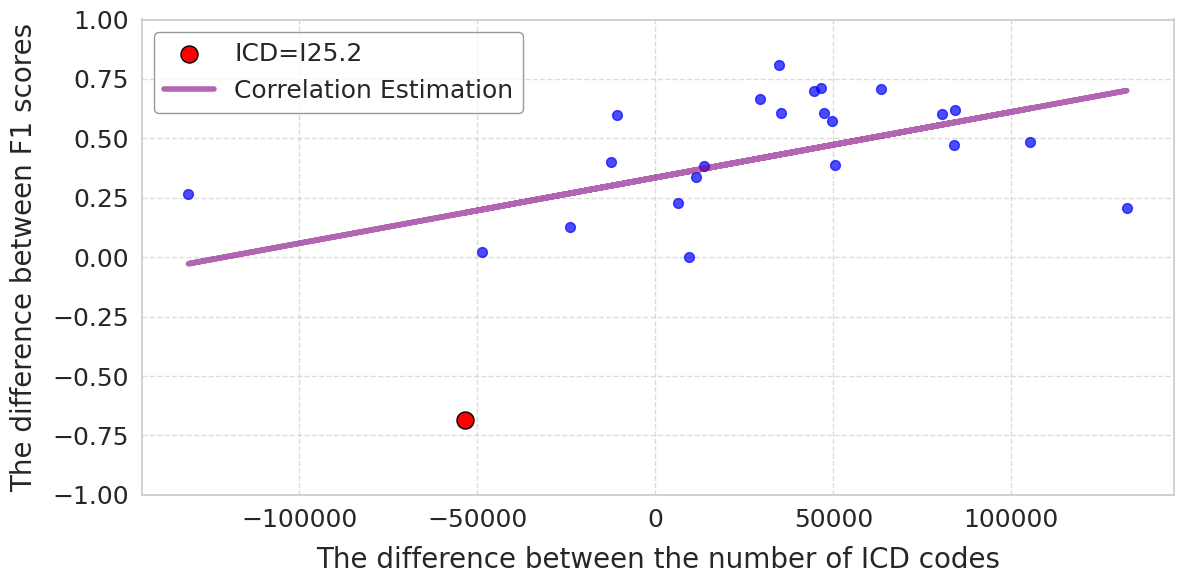} 
    \caption{Dependency between differences in the number of codes in original and linked train sets and corresponding F1 scores differences on the common test.}
    \label{fig:delta_scatter} 
   \end{figure}
    \begin{figure}[t!]
    \vspace{-0.2cm}
        \centering
    \includegraphics[width=0.45\textwidth]{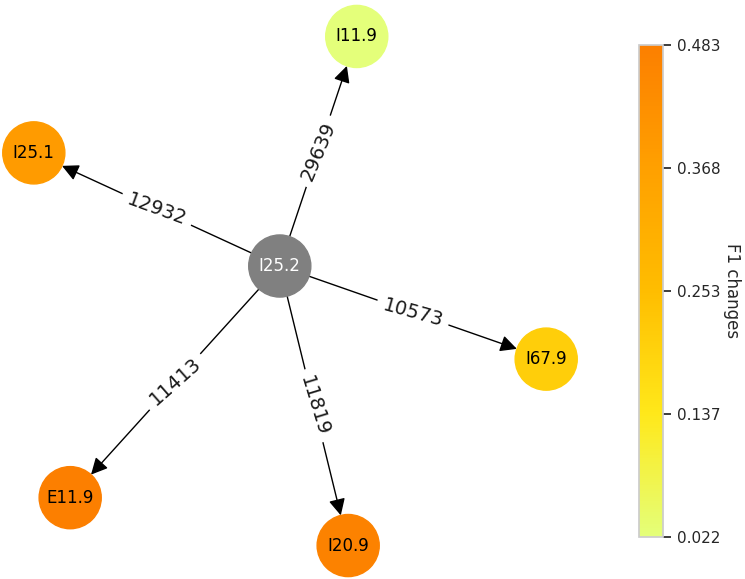} 
    \caption{The relationship between transitions from I25.2 and F1 improvements: numbers on the arrows indicate transition frequency, while node color intensity represents the magnitude of F1 metric change.}
    \label{fig:graph} 
    \vspace{-0.2cm}
\end{figure}

\subsection{Disease-Wise Quality Shift Analysis}


\paragraph{Linked Data's Improvement Stability} Fig.~\ref{fig:delta_scatter} shows how changes in appointment counts from \textit{original} to \textit{linked} data affect the diagnosing F1 score. Notably, F1 scores generally improved for the majority of ICD codes regardless of appointment counts increase or decrease. Although the performance change varies, the observations suggest the improved class balance. The only exception is the \textit{I25.2 Past myocardial infarction} code, which will be analyzed in detail in the subsequent section.

\paragraph{Case Study: Diagnosing Degradation}
In Fig.~\ref{fig:delta_scatter}, a sharp F1 score drop is observed for \textit{I25.2 Past myocardial infarction}. Apparently, the disease has been mistakingly re-linked to other errounessly. We studied the case by analyzing which ICD codes has \textit{I25.2} been replaced with.
Fig.~\ref{fig:graph} shows the most frequent transition (I25.2 to I11.9, \textit{Hypertensive heart disease}) yielded minimal F1 improvement (0.02), likely due to symptom overlap. E11.9 (\textit{Type 2 diabetes mellitus}) had the highest gain (0.48) due to clearer distinctions. I25.1 (\textit{Atherosclerotic heart disease}) and I20.9 (\textit{Angina pectoris}) had significant gains (0.38, 0.47), while I67.9 (\textit{Unspecified cerebrovascular disease}) had a moderate gain (0.21).
From the results, clearly distinguishable diagnoses yield higher F1 scores compared to those with symptom overlap.

\section{Conclusion}

In this paper, we presented the first models for multi-label ICD-10 coding of electronic health records (EHRs) in Russian. Our study focuses on two key tasks: information extraction from the diagnosis field of EHRs and diagnosis prediction based on a patient's medical history. The NLP pipeline developed for the first task was utilized to re-annotate EHRs in the training set for the second task. The results demonstrate that fine-tuned LMs significantly enhance performance in predicting ICD codes from past medical history. Specifically, the model trained on automatically linked data exhibited faster learning and better generalization compared to the original dataset, achieving higher weighted F1 scores early in training, while the original model plateaued with minimal improvements.
Notably, the linked data model consistently outperformed the original across both frequent and rare ICD classes, achieving higher F1 scores together with improved confusion metrics (TP/FP/FN/TN) and reduced variability. These results suggest that the linked dataset enables effective handling of both common and rare ICD codes. Overall, our findings highlight the importance of a neural pipeline for automating ICD coding and improving the accuracy and informativeness of medical text labeling.

Future research will focus on the integration of additional external medical sources like knowledge graphs to improve ICD code prediction. We plan to study the generalization of LLMs on rare codes. 


\newpage

\section*{Limitations}
\paragraph{Other biomedical corpora in Russian} The most relevant corpora to our study are RuCCoN \cite{nesterov-etal-2022-ruccon} and NEREL-BIO \cite{NERELBIO,loukachevitch-etal-2024-biomedical}, which link entities from clinical records or PubMed abstracts to the Russian part of UMLS.
We performed preliminary transfer learning experiments with these two corpora, but a thorough analysis of the semantic differences across all three corpora is future work.


\paragraph{Clinical Diversity} Our dataset  may not fully capture the diversity of clinical scenarios and patient demographics. A more varied dataset could improve the robustness and generalizability of the models. Clinical language can vary significantly across different medical specialties and institutions. This variability may impact the model's ability to generalize across various clinical contexts.

\paragraph{Data Imbalance} The dataset may suffer from class imbalance, with certain ICD codes being underrepresented. This could affect a model's ability to accurately predict less common diagnoses.

\section*{Ethics Statement}

\paragraph{No Personal Patient Data in \dataset{}} 
\dataset{} does not contain any personally identifiable patient information. The dataset consists solely of diagnosis conclusions written by medical professionals, which were manually labeled based on the ICD-10 CM (Clinical Modification) system. Prior to the annotation process, annotators were instructed to ensure that no personal information was included in the conclusions. Their task was to identify and remove any personal or identifiable information manually from these texts. No patient-related information is disclosed in \dataset{}.

\paragraph{Private in-house EHR data in \datasettt{}} Diagnosis prediction leverages prior EHR data along with details from the current visit. As a source for \datasettt{}, we utilize records from the Medical Information System of a major European city. All patients, prior to visiting a doctor, sign a special consent form for the processing of their data. The EHR data, which forms the foundation of \dataset{}, is an in-house dataset that will not be released.

\paragraph{Human Annotations}
All data annotations in this paper are newly created.
Dataset annotation was conducted by annotators, and there are no associated concerns (e.g. regarding compensation). Each annotator was paid a rate of \$12 per hour for their contributions. An estimated 85 hours of annotation work per expert resulted in a total payment of \$1,020 per annotator.
Russia’s full-time monthly minimum wage is under \$200, highlighting the substantial effort and investment in creating this high-quality resource.
All annotators were aware of potential annotation usage for research purposes.

\paragraph{Inference Costs} Running the complete evaluation experiment on a single V100 GPU takes approximately 7.5h and 11h for a decoder-only and encoder-only LM, respectively, while the LLM with RAG evaluation experiment on a single A100 GPU takes approximately 5.5h.

\paragraph{Potential Misuse} The RuCCoD dataset, intended for ICD coding in Russian, may be misused if not handled correctly. Potential issues include inaccurate applications leading to incorrect code assignments and overreliance on automated systems without proper validation. To prevent these problems, it is crucial to provide clear guidelines and adequate training for doctors on using AI assistants, ensuring compliance with ethical and legal standards in research and healthcare. 

\paragraph{Transparency} The \dataset{} and all associated annotation materials are being released under the CC BY 4.0 license. It should be noted that the dataset contains only diagnosis codes and no medical histories or personal patient data. Furthermore, all diagnoses have been rigorously verified to ensure complete anonymity, in accordance with the prevailing norms of open research practice. 

\paragraph{Use of AI Assistants} We used Grammarly to proofread the paper, correcting grammatical, spelling, and stylistic errors, as well as rephrasing sentences. Consequently, certain sections of our paper may be identified as AI-generated, AI-edited, or a combination of human and AI contributions.

\section*{Acknowledgements}
This work was supported by a grant, provided by the Ministry of Economic Development of the Russian Federation in accordance with the subsidy agreement (agreement identifier 000000C313925P4G0002) and the agreement with the Ivannikov Institute for System Programming of the Russian Academy of Sciences dated June 20, 2025 No. 139-15-2025-011.

\bibliography{custom, anthology}

\appendix

\section{Appendix: Annotation Details}\label{app:ann_gui}
\subsection{Task Overview}
The task is to review the diagnoses in the BRAT markup system, categorize them into separate entities corresponding to individual nosologic units, and assign each of the selected entities an identifier in the form of an ICD code from the provided clinical modification of the ICD-10-CM classifier. The purpose of the annotation is to assign the correct, most private (to the extent possible from the limited anamnesis cotext) identifier to each nosologic unit represented in the diagnosis.  

\subsection{Data and resources}
\textit{Data}. The documents you will be annotating are anonymized diagnoses. To facilitate and speed up the annotation process, most nosologic units are highlighted and pre-labeled with an ICD code. 

\textit{Vocabulary}. Each phrase identified in the text as a nosological unit or not highlighted but being such must be associated with a code from the ICD-10. This markup will use the clinical modification of the ICD-10-CM, which includes about 17762 different medical diagnoses. 

\textit{Additional Resources}. Although the markup system is already loaded with the ICD-10, you can use the following additional resources to help you correctly identify the most appropriate ICD code:
\begin{itemize}
    \item The ICD Code Clinical Modification Version 10 is a Russian-language web service for searching and determining the optimal ICD code, available at: www.mkb-10.com. Registration is not required to access this resource.
    \item Google - You can use Google if you are unfamiliar with a clinical diagnosis or if you encounter a previously unknown abbreviation or acronym.
    \item Wikipedia - You can also use Wikipedia to find additional information.
\end{itemize}

\subsection{Task Description}
For each selected or unselected piece of text corresponding to a nosological unit, you need to assign an ICD code.
\textit{Example}: “Atopic dermatitis in partial remission disseminated form”. The selected text fragment “Atopic dermatitis in partial remission” should be associated with the diagnosis "Other atopic dermatitis" (L20.8). 
 Make sure that no text fragment representing a nosological unit is left without an assigned ICD code, thus ensuring the completeness of the markup.

 However, each nosologic unit should correspond to only one code. However, in many cases, the selected nosologic units may correspond to more than one ICD code, in which case you should follow the following rules:
 \begin{itemize}
     \item 1. Select an ICD code that maximizes the specificity of the diagnosis up to subsection X.00. 
     \item 2. If the nosological unit includes modifiers such as “mild”, “severe”, “acute”, “chronic”, indication of degree, stage, etc., the modifier should be taken into account when searching for the appropriate ICD code. However, it is often the case that the classifier will only have a more general diagnosis that does not include the above modifier. In this case, select the optimal ICD code by ignoring the modifier. However, modifiers that are inseparable in meaning from the underlying concept should always be considered when selecting the optimal ICD code (e.g., “Acute myocardial ischemia”).
 \end{itemize}

The following rules should also be followed when marking up:
\begin{itemize}
    \item If the selected nosological unit is written in the plural and the corresponding ICD code exists in the classifier in the plural, you should select it. Otherwise, you should search for the ICD code in the singular.
    \item Sometimes in the classifier there are diagnoses that at first glance seem to be absolutely identical, which can be differentiated only by the context of the electronic medical record. 
\end{itemize}

\subsection{Annotation Tool}
The annotation process is conducted using a specialized web service called brat (https://brat.nlplab.org/). You will be provided with a customized login and password. All necessary information from clinical diagnoses and preliminary markup with ICD codes are entered into the annotation tool. Each document in the brat web service leads to a separate clinical diagnosis. 

Each selected text fragment is a nosological unit to be associated with the corresponding ICD code. In order to call the ICD code selection menu, you need to highlight the section of text you are going to mark up or double-click on the green label “icd\_code” located above the selected text fragment. If you think that a section of the diagnosis is selected incorrectly or redundantly, you need to correct or delete the corresponding selection.  

The window may or may not have a pre-selected ICD code on the Ref line. If specified, compare the correctness of the ICD code specified in the “Ref” line with the selected text fragment specified in the “Text” field. If the ICD code is correct, press the “OK” button and move to the next selected text fragment. If the ICD-code is not specified or is specified incorrectly, double-click the “Ref” line in the “Normalization” field, and the ICD-code search window will open. In the opened window check the correctness of the diagnosis selection for search in the “Query” line and click on the “Search ICD\_codes” button. The system will search in the ICD codes classifier and list them. If the system does not find the codes by the specified text fragment, try to change it. 

Select the appropriate ICD code and its decoding from the list and press the “OK” button (or double-click on the required ICD code). The system will save your selection and return to the previous window, where you should also click on the “OK” button. The system will remember your selection and you can proceed to annotate the next selected text section. 

If you did not find a suitable ICD code in the list of ICD codes found by the system, you can try to change the search phrase in the “Query” field, by which the search is performed, and perform the search again. In most cases, the correct selection of the search phrase allows one to find the most appropriate ICD code in the classifier.

If the built-in search system does not yield results, you can switch to the external directory of ICD codes specified in A2. To do this, click on the magnifying glass icon in the “Normalization” field. You can also go to the Google search engine and Wikipedia web encyclopedia by clicking on the corresponding link in the “Search” field.   

If even after changing the search phrase and searching in external resources you cannot find a suitable ICD code, return to the previous menu by clicking on the “cancel” button and delete the identifier located in the ID line in the “Normalization” field in the opened window. The same should be done if a text section that is not a nosological unit is selected. Deleting the identifier will clear the “Ref” line; this will serve as an indicator that the selected text fragment could not be matched with a suitable ICD code.

\subsection{Inter-Annotator Agreement}
To assess consistency among experts manual ICD coding, we first measured \textit{Inter-Annotator Agreement} (IAA) metric, defined as the ratio of accepted codes to the total number of unique codes assigned per record~\cite{LUO2019103132}. The overall IAA reached 50\%, indicating a moderate level of agreement. Such values align with previous reports on ICD coding, where Kappa coefficients typically range from 27\% to 42\% (corresponding to agreement rates of 29.2\%–46.8\%)~\cite{STAUSBERG200850}, and coding accuracy varies widely from 41.8\% to 98\% depending on domain and code granularity~\cite{campbell2001systematic,hosseini2021factors}. These observations confirm that moderate agreement is expected given the inherent complexity and subjectivity of assigning fine-grained ICD codes.

To further ensure the consistency of obtained annotations and provide an additional comparative assessment of agreement, we also measured agreement using the \textit{Jaccard similarity coefficient} (JSC). This metric captures the ratio of the intersection to the union of ICD code sets assigned by two annotators and is particularly suited for multi-label settings where overlapping but not identical code sets are common. The mean pairwise JSC values were 0.308, 0.336, and 0.366 for annotator pairs (1–2), (1–3), and (2–3), respectively. These results are consistent with existing studies on ICD coding from unstructured clinical text, which report average Jaccard agreements ranging from 0.179 to 0.345 and, for certain code categories, scores approaching zero~\cite{xu2019multimodal,skevo2020sr02}.

Together, the IAA and JSC analyses indicate that the observed annotation variability is realistic and reflects the intrinsic difficulty of achieving high inter-expert consistency in multi-label ICD coding.


\section{Related Work}\label{appx:rel}
In describing our work, we encountered persistent terminological ambiguity arising from overlapping nomenclature for distinct task formulations. For instance, the term “ICD coding” is broadly applied to both (1) multi-label classification of medical texts (e.g., assigning ICD codes to discharge summaries) \cite{li2020icd, vu2020label, wang2024icdxml} and (2) entity linking, where discrete clinical diagnoses are mapped to specific codes \cite{lavergne2016dataset, neveol2017clef, coutinho2022transformer}. This conflation obscures fundamental differences: the former treats coding as document-level prediction to capture all relevant codes for a patient’s condition, while the latter focuses on precise alignment of clinical entities (e.g., distinguishing “acute myocardial infarction” from its subtypes) through semantic matching, addressing challenges like synonymy or hierarchical code relationships. To resolve this ambiguity, in our work we propose explicit terminology: \textbf{“ICD coding”} refers to multi-label classification of medical texts, whereas \textbf{“Medical entity linking”} denotes entity-level code assignment.

\paragraph{ICD coding} ICD coding has traditionally relied on established machine learning techniques. Early approaches employed methods such as Support Vector Machines (SVM) with TF-IDF features to represent clinical notes \cite{perotte2014diagnosis}. Feature engineering, including gradient boosting for large datasets, also played a significant role in enhancing ICD coding accuracy \cite{diao2021automated}. Regular expression-based mapping and adaptive data processing further improved efficiency in specific healthcare settings \cite{zhou2020construction}.

The advent of neural networks marked a paradigm shift in ICD coding. Recurrent Neural Networks (RNNs), including LSTMs and GRUs, were utilized to encode EHR data and capture temporal dependencies within clinical notes \cite{choi2016doctor, Baumel2018MultiLabelCO}. Convolutional Neural Networks (CNNs) offered alternative architectures for extracting features from clinical text, with models like CAML demonstrating their effectiveness \cite{mullenbach-etal-2018-explainable}. Subsequent advancements introduced multi-filter CNNs \cite{li2020icd} and squeeze-and-excitation networks in CNN \cite{liu2021effective} to enhance feature extraction. Addressing the challenge of imbalanced code distribution, researchers introduced focal loss \cite{liu2021effective} and self-distillation mechanisms to improve prediction accuracy for rare codes \cite{zhou2021automatic}. Other models, like HA-GRUs used the charachter-level information \cite{Baumel2018MultiLabelCO}. Ensemble models used CNN, LSTM, and decision trees to improve accuracy \cite{Xu2018MultimodalML}.

A crucial line of research has focused on integrating external medical knowledge and the inherent hierarchical structure of ICD codes. Approaches have incorporated medical definitions \cite{DBLP:journals/corr/abs-1711-04075}, Wikipedia data for rare diseases \cite{Bai2019ImprovingMC} and medical ontologies \cite{Bao2021MedicalCP} to enrich term embeddings. Tree-of-sequences LSTMs \cite{xie-xing-2018-neural} and graph neural networks \cite{cao2020hypercore, Xie2019EHRCW} were developed to capture relationships between codes, either through hierarchical structures or co-occurrence patterns. Models like KG-MultiResCNN leveraged external knowledge for relations understanding \cite{boukhers2023knowledge}. Weak supervision was used to overcome the lack of training data \cite{dong2021rare, gao2022classifying}. Furthermore, domain-specific pre-trained language models (PLMs) such as BioBERT \cite{10.1093/bioinformatics/btz682}, ClinicalBERT \cite{Alsentzer2019PubliclyAC}, and PubMedBERT \cite{gu2021domain} have shown promise in improving performance on various biomedical tasks. However, adapting these models to the large-scale, multi-label nature of ICD coding presents unique challenges, particularly regarding long input sequences \cite{pascual2021towards, ji2021does}. Recent efforts, such as BERT-XML \cite{zhang2020bert}, have addressed this through input splitting and label attention mechanisms. Read, Attend, and Code (RAC) was proposed by Kim and Ganapathi \cite{kim2021read} and achieved state-of-the-art results. Despite these developments, challenges remain in handling semi-structured text and variability of notes \cite{lu2023towards}.

Recent studies have increasingly focused on leveraging attention mechanisms and improving the interaction between clinical note representations and ICD code representations. Models such as LAAT \cite{vu2020label} and EffectiveCAN \cite{liu2021effective} have incorporated refined label-aware attention mechanisms. However, the effective application of PLMs to ICD coding requires careful consideration of input length constraints and the development of robust mechanisms for capturing long-range dependencies. Also, the models need to better understand relationships between different sections of clinical notes \cite{lu2023towards}.

\paragraph{Diagnosis prediction} Diagnosis prediction using structured EHR data has been extensively studied with deep learning approaches. NECHO~\cite{koo2024necho} improves next-visit diagnosis prediction by centering learning on medical codes and incorporating hierarchical regularization to capture structured dependencies in EHR data. DPSS~\cite{zhang2020dpss} enhances predictive robustness by modeling patient records as sequences of unordered clinical events, preserving temporal patterns while mitigating biases introduced by the artificial ordering of medical records. The importance of patient history in EHR-based diagnosis prediction demonstrates that historical records alone can achieve 76.6\% accuracy, which increases to 93.3\% when structured physical examination and laboratory data are integrated~\cite{fukuzawa2024history}. At the population level, applying a Bi-GRU model trained on structured EHR data with SNOMED embeddings to predict chronic disease onset demonstrates the utility of structured clinical histories in early disease identification~\cite{grout2024predicting}. To optimize the use of structured medical codes for diagnosis prediction, MERA \cite{ma2025mera} introduces hierarchical contrastive learning and ranking mechanisms to refine diagnosis classification within large ICD code spaces. These studies collectively illustrate the evolution of EHR-based diagnosis prediction from sequence modeling to hierarchical representation learning, highlighting the role of structured clinical history in improving predictive accuracy.

\paragraph{RAG} LLMs face challenges as standalone systems for high-precision tasks such as ICD-linking, primarily due to their limited accuracy in extracting detailed, domain-specific information. Ma et al.\cite{Ma_2023} demonstrated that while LLMs lag behind fine-tuned SLMs in information extraction tasks, they excel in understanding and reorganizing semantic content, making them effective at reranking retrieved information. To overcome the limitations of accuracy and domain specificity, recent approaches have incorporated Retrieval-Augmented Generation (RAG) techniques. RAG combines the structured knowledge of external databases for retrieval with the semantic reasoning strengths of LLMs for reranking, resulting in improved precision and overall task performance.

\citet{Klang2024.10.15.24315526} demonstrated the effectiveness of RAG in enhancing LLMs for ICD-10-CM medical coding. Their study revealed that RAG-enhanced LLMs outperform human coders in accuracy and specificity, emphasizing the potential of retrieval mechanisms in improving clinical documentation. Similarly, \citet{kwan2024largelanguagemodelsgood} proposed a two-stage Retrieve-Rank system for medical coding, achieving a perfect match rate for ICD-10-CM codes and significantly surpassing vanilla LLMs.
The MedCodER framework \cite{baksi2024medcodergenerativeaiassistant} leverages a pipeline of extraction, retrieval, and reranking, to improve automation and interpretability in ICD-10 coding. It demonstrates SOTA performance on ACI-BENCH by integrating LLMs with semantic search and evidence-based reasoning. Boyle et al. \cite{boyle2023automatedclinicalcodingusing} presented a zero-shot ICD coding approach using LLMs and a tree-search strategy, achieving a SOTA on the CodiEsp dataset, particularly excelling in rare code prediction without task-specific training.
Abdulnazar et al. \cite{10703053} applied GPT-4 for clinical text cleansing to enhance MCN. By combining text standardization with RAG, their method improved mapping precision to SNOMED CT in the German language.

\begin{table*}[t!]
\centering
\scalebox{0.83}{{\begin{tabular}{ccccccc}
\toprule
  \textbf{Task} & \textbf{Model or Approach} & \textbf{LR} &  \textbf{\# Epochs} &  \textbf{BS} & \textbf{Scheduler} &\textbf{WD} \\
\midrule
NER & RuBioBERT & 1e-5 & 20 & 32 & Cosine~\cite{loshchilov2017sgdrstochasticgradientdescent} & 0.01 \\ 
EL & BERGAMOT+BioSyn & 2e-5 & 20 & 32 & Adam~\citep{DBLP:journals/corr/KingmaB14} & 0.01\\ 
LLM tuning & LoRA & 5e-5 & 33 & 2 & Linear with Warmup & 0.01\\
ICD code prediction & Longformer & 5e-5 & 2 & 4  & Linear with Warmup & 0.01\\
\bottomrule
\end{tabular}
}}\caption{\label{tab:hp} Models and training hyperparameters. LR stands for learning rate, BS for batch size, WD for weight decay}\end{table*}

\section{BERT-based NER Results}\label{appx:ner-bert}

Tab. \ref{tab:ner-results} presents evaluation results for NER task on the \dataset{} dataset. In the context of NER, RuBioBERT employs a softmax activation function in its output layer. BINDER utilizes RuBioBERT backbone and approaches NER as a representation learning problem by maximizing the similarity between the vector representations of an entity mention and its corresponding type \cite{zhangoptimizing}. RuBioBERT achieves the highest F1-score of 0.756 when trained on the \dataset{}, suggesting that this dataset is particularly effective for the model. BINDER trained on \dataset{} achieves an F1-score of 0.71, slightly lower than RuBioBERT trained on the same dataset.


\begin{table*}[t!]
\centering
\begin{tabular}{l|llll}
Model & \textbf{Train Data} & \textbf{F1-score} & \textbf{Precision} & \textbf{Recall} \\\midrule
RuBioBERT & \dataset{} train & 0.756 & 0.75 & 0.77 \\
RuBioBERT & BIO-NNE train & 0.62 & 0.57 & 0.67 \\
RuBioBERT & \dataset{} + BioNNE train & 0.72 & 0.75 & 0.70 \\
BINDER + RuBioBERT & \dataset{} train & 0.71 & 0.72 & 0.71 \\\bottomrule
\end{tabular}
\caption{Evaluation results for NER task on \dataset{} dataset.}
\label{tab:ner-results}
\end{table*}

\section{Entity Linking Results}\label{appx:cross_terminology_transfer} 

\begin{table*}[t!]
\centering

\begin{tabular}{l|cccccc}

\multirow{2}{*}{\textbf{Train set}}    & 
\multicolumn{2}{c}{\textbf{SapBERT}}  & \multicolumn{2}{c}{\textbf{CODER}} & \multicolumn{2}{c}{\textbf{BERGAMOT}}  \\
\cmidrule(lr){2-3} 
\cmidrule(lr){4-5}
\cmidrule(lr){6-7}
   &  @1 & @5 & @1 & @5 & @1 & @5 \\  
\midrule 
\multicolumn{7}{c}{\textbf{Zero-shot evaluation, strict}} \\
\midrule
ICD dict & 0.3327 & \textbf{0.5712} & 0.2631 & 0.4687 & 0.3495 & \textbf{0.6170}  \\
ICD dict+UMLS synonyms & \textbf{0.3546} & 0.5197 & \textbf{0.3237} & \textbf{0.4765} & \textbf{0.3559} & 0.5487 \\
\midrule 
\multicolumn{7}{c}{\textbf{Supervised evaluation, strict}} \\
\midrule
ICD & \textbf{0.6132} & \textbf{0.8182} & \textbf{0.6202} & \textbf{0.8169} & \textbf{0.6415} & \textbf{0.8459} \\
ICD+UMLS sumonyms &  0.5326 & 0.7382 & 0.5358 & 0.7318 & 0.4984 & 0.7253 \\
RuCCoN  & 0.3591 & 0.5345 & 0.3598 & 0.5732 & 0.3643 & 0.5313 \\
RuCCoN+ICD  &  0.3952 & 0.5732 & 0.3888 & 0.6570 & 0.3817 & 0.5983 \\
NEREL-BIO  &  0.3443 & 0.4913 & 0.3378 & 0.5274 & 0.3353 & 0.5113 \\
NEREL-BIO+ICD  & 0.3804 & 0.5596 & 0.3804 & 0.6325 & 0.3598 & 0.5525 \\
\midrule 
\multicolumn{7}{c}{\textbf{Zero-shot evaluation, relaxed}} \\\midrule
ICD dict & 0.4842 & \textbf{0.6886} & 0.3752 & 0.6190 & 0.5035 & \textbf{0.7286}   \\
ICD dict+UMLS synonyms  & \textbf{0.5551} & 0.6867 & \textbf{0.5055} & \textbf{0.6293} & \textbf{0.5603} & 0.7073 \\
\midrule 
\multicolumn{7}{c}{\textbf{Supervised evaluation, relaxed}} \\\midrule
ICD   & 0.7763 & \textbf{0.8839} & \textbf{0.7872} & 0.8743 & \textbf{0.7917} & \textbf{0.8943} \\ 
ICD+UMLS sumonyms  & \textbf{0.7788} & 0.8616 & 0.7714 & \textbf{0.8860} & 0.7449 & 0.8738 \\ 
RuCCoN  & 0.5235 & 0.6531 & 0.5429 & 0.7208 & 0.5132 & 0.6564 \\
RuCCoN+ICD     & 0.5493 & 0.6602 & 0.5770 & 0.7485 & 0.5571 & 0.6873 \\
NEREL-BIO     & 0.4803 & 0.6067 & 0.4958 & 0.6634 & 0.4778 & 0.6170 \\
NEREL-BIO+ICD   & 0.5455 & 0.6447 & 0.5474 & 0.7292 & 0.5384 & 0.6505 \\
\bottomrule
\end{tabular}
\caption{Cross-domain transfer results for biomedical linking models. Evaluation results for linking models trained on RuCOD, RuCCoN, NEREL-BIO as well as their union.  \emph{ICD+UMLS synonyms} stands for ICD train set with the vocabulary enriched with ICD disease name synonyms from the UMLS knowledge base. The best results for each model and set-up are highlighted in \textbf{bold}. }\label{tab:res_entity_linking_transfer}


\end{table*}

Since there are many datasets for entity linking in the biomedical domain, including corpora in Russian, we explored whether these corpora can be helpful for ICD coding. Additionally, we attempted to enrich the ICD normalization vocabulary with concept names from the Unified Medical Language System (UMLS) metathesaurus which includes the ICD-10 vocabulary. Specifically, for each ICD code, we find its Concept Unique Identifier (CUI) in UMLS and retrieve all concept names that share the same CUI but are adopted from the source vocabularies different from ICD-10. We employ the following Russian biomedical corpora for experiments on cross-terminology transfer:

\emph{RuCCoN}~\citep{nesterov-etal-2022-ruccon} is a manually annotated corpus of clinical records in Russian. It contains 16,028 mentions linked to 2,409 unique concepts from the Russian subset of UMLS metathesaurus~\citep{bodenreider2004unified}.

\emph{NEREL-BIO}~\citep{NERELBIO,loukachevitch-etal-2024-biomedical} is a corpus of 756 PubMed abstracts in Russian manually linked to 4,544 unique UMLS concepts. The corpus is specifically focused on two main problems: (i) entity nestedness and (ii) cross-lingual Russian-to-English normalization for the incomplete Russian UMLS terminology. In total, NEREL-BIO provides 23,641 entity mentions manually linked to 4,544 unique UMLS concepts. 4,424 mentions have no concept name representation in the Russian UMLS subset and are linked to 1,535 unique concepts present in the English UMLS only.

We experiment with three state-of-the-art specialized biomedical entity linking models:

 \emph{SapBERT} is a metric learning framework that learns from synonymous UMLS concept names by generating hard triplets for pre-training~\citep{liu-etal-2021-self,liu-etal-2021-learning-domain}.
 
 \emph{CODER} is a contrastive learning model inspired by semantic matching methods that use both synonyms and relations from the UMLS~\citep{CODER-BERT} to learn concept representations.
 
 \emph{BERGAMOT} is an extension of SapBERT which learns concept name-based and graph-based concept representations simultaneously and introduces a cross-modal alignment loss to transfer knowledge from a graph encoder to a BERT-based language encoder~\citep{sakhovskiy-etal-2024-biomedical}. The graph encoder is discarded after the pretraining stage and only a BERT encoder is used for inference.

For supervised entity linking, we adopt BioSyn~\citep{sung-etal-2020-biomedical}, a BERT-based framework that iteratively updates entity representations using synonym marginalization. For each dataset, we trained BioSyn with default hyperparameters for 20 epochs.

\paragraph{Relaxed EL Evaluation} We assess two entity linking set-ups: (i) \textbf{strict} evaluation which implies an exact match between predicted and ground truth codes and (ii) \textbf{relaxed} evaluation with all codes being truncated to 3-symbols codes (corresponding to the second level of hierarchy).

The results of cross-terminology entity linking transfer presented in Tab.~\ref{tab:res_entity_linking_transfer} and Tab.~\ref{tab:ruccod_ie} reveal a few insightful findings related to linking ICD codes.

\paragraph{Vocabulary Extension is not a Cure} While extension of ICD vocabulary consistently gives a slightly improved Accuracy@1 in a zero-shot setting, additional synonyms introduce severe noise in a supervised setting. Specifically, a significant drop of 8.1\%, 8.4\%, 14.3\% Accuracy@1 is observed for SapBERT, CODER, and BERGAMOT, respectively. Even in an unsupervised setting, vocabulary extension drops Accuracy@5 by 5.2\% and 6.8\% for SapBERT and BERGAMOT, respectively.

\paragraph{Complicated Cross-Terminology Transfer} Both training on RuCCoN and NEREL-BIO as well merge of these corpora with \dataset{} do not lead to improvement over zero-shot coding. The finding indicates the specificity and high complexity of ICD coding within the entity linking task.

\paragraph{Complexity of Fine-Grained ICD coding} The high gap between the strict and supervised evaluation of around 15\% Accuracy@1 indicates that distinguishing between semantically similar diseases sharing the same therapeutic group is a major challenge. 

\section{LLM with RAG results}\label{app:llm_rag_res}
All LLM with RAG experiments were conducted with a temperature setting of 0 for all LLMs and a top-k value of 15 for the number of retrieved entities from similarity search. The LLMs used are specified in Appx. \ref{appx:impl_details}. For the embedding model, we utilized BERGAMOT. To construct the vector database, we used dictionaries extracted from NEREL-BIO, RuCCoN, the ICD dictionary, and the ICD dictionary combined with \dataset. The results are presented in Tables \ref{rag_res_1} and \ref{rag_res_2} for strict evaluation, and in Tables \ref{rag_res_3} and \ref{rag_res_4} for relaxed evaluation.

For the NER task, the ICD dict.+\dataset{} dataset yielded the best results. The Llama3.1:8b-instruct-fp16 model achieved the highest F-score (0.511), precision (0.580), recall (0.456), and accuracy (0.343). Qwen2.5-7B-Instruct and Llama3-Med42-8B followed with F-scores of 0.495 and 0.491, respectively. In contrast, NEREL-BIO and RuCCoN datasets showed significantly lower performance, with F-scores below 0.13 and accuracies under 0.07.

For NER+ICD Linking, the same dataset and model led again, with Llama3.1:8b-instruct-fp16 achieving an F-score of 0.268 and accuracy of 0.155. Qwen2.5-7B-Instruct and Llama3-Med42-8B followed closely with F-scores around 0.245. Performance on NEREL-BIO and RuCCoN was much lower, with F-scores under 0.022 and accuracies below 0.011.

For ICD Code assignment, Llama3.1:8b-instruct-fp16 also performed best, with an F-score of 0.458 and accuracy of 0.297. Qwen2.5-7B-Instruct and Llama3-Med42-8B also performed well, with F-scores of 0.463 and 0.457. Again, NEREL-BIO and RuCCoN datasets exhibited weaker results, with F-scores below 0.15 and accuracies under 0.09.

In summary, the ICD dict.+\dataset{} dataset consistently outperformed others with Llama3.1:8b-instruct-fp16 being the best model. Relaxed evaluation settings produced similar trends.


\section{LLM with tuning results}\label{app:llm_tuning_res}
The LLM tuning results are in Tab. \ref{tab:llm_tune}.

For the NER task, Llama3-Med42-8B achieved the highest F-score of 0.642, which corresponds to the highest Precision and Recall among the models. Phi3\_5\_mini and Mistral-Nemo demonstrated similar performance (F-scores of 0.627 and 0.614, respectively), but slightly lag behind the leader. The Qwen2.5-7B-Instruct model showed the lowest scores across all metrics, with an F-score of 0.565 and an Accuracy of 0.393.

In the NER + ICD linking task, the use of the \dataset{} or BERGAMOT approach significantly improved the linking performance. For instance, Phi3\_5\_mini achieved the highest F-score of 0.333 when using \dataset{}, and Llama3-Med42-8B reached an F-score of 0.299. Notably, for all models, the use of \dataset{} proved to be more beneficial than the BERGAMOT approach.

In the ICD code assignment task, results also improved significantly with the use of the \dataset{} dataset. Once again, Phi3\_5\_mini emerged as the top-performing model, attaining an F-score of 0.480 when using \dataset{}. Llama3-Med42-8B and Mistral-Nemo also demonstrated strong results, with F-scores of 0.435 and 0.446, respectively, when using \dataset{}. It is noteworthy that the inclusion of \dataset{} consistently improved Precision and Recall across all models.

Based on the presented results, it can be concluded that for all tasks (NER, NER+Linking, and ICD code assignment), the use of \dataset{} significantly enhances model performance compared to relying solely on the dictionary or embeddings. The top-performing models across all tasks are Llama3-Med42-8B and Phi3\_5\_mini, indicating their high efficiency in medical tasks following PEFT tuning.

\begin{table*}[b!]\label{llm_tune_table}
\centering
\begin{tabular}{lrrrr}

\toprule
\multirow{1}{*}{\textbf{Model}} & \multicolumn{1}{c}{\textbf{Precision}}  & \multicolumn{1}{c}{\textbf{Recall}} & \multicolumn{1}{c}{\textbf{F-score}} & \multicolumn{1}{c}{\textbf{Accuracy}}  \\
\midrule 
\multicolumn{5}{c}{\textbf{NER}} \\\midrule
Llama3-Med42-8B, \dataset{} & \textbf{0.642} & \textbf{0.642} & \textbf{0.642} & \textbf{0.473} \\
Qwen2.5-7B-Instruct, \dataset{} & 0.567 & 0.562 & 0.565 & 0.393 \\
Phi3\_5\_mini, \dataset{} & 0.632 & 0.623 & 0.627 & 0.457 \\
Mistral-Nemo, \dataset{} & 0.631 & 0.598 & 0.614 & 0.443 \\

\midrule
\multicolumn{5}{c}{\textbf{NER+Linking}} \\\midrule
Llama3-Med42-8B, ICD dict. & 0.149 & 0.149 & 0.149 & 0.08 \\
Llama3-Med42-8B, ICD dict. + \dataset{} & 0.299 & 0.299 & 0.299 & 0.176 \\
Llama3-Med42-8B, ICD dict. + BERGAMOT & 0.286 & 0.286 & 0.286 & 0.167 \\
Qwen2.5-7B-Instruct, ICD dict. & 0.188 & 0.186 & 0.187 & 0.103 \\
Qwen2.5-7B-Instruct, ICD dict. + \dataset{} & 0.281 & 0.279 & 0.28 & 0.163 \\
Qwen2.5-7B-Instruct, ICD dict. + BERGAMOT & 0.2 & 0.198 & 0.199 & 0.11 \\
Phi3\_5\_mini, ICD dict. & 0.272 & 0.268 & 0.27 & 0.156 \\
Phi3\_5\_mini, ICD dict. + \dataset{} & \textbf{0.335} & \textbf{0.33} & \textbf{0.333} & \textbf{0.199} \\
Phi3\_5\_mini, ICD dict. + BERGAMOT & 0.322 & 0.317 & 0.32 & 0.19 \\
Mistral-Nemo, ICD dict. & 0.231 & 0.219 & 0.224 & 0.126 \\
Mistral-Nemo, ICD dict. + \dataset{} & 0.303 & 0.287 & 0.295 & 0.173 \\
Mistral-Nemo, ICD dict. + BERGAMOT & 0.267 & 0.253 & 0.26 & 0.149 \\

\midrule
\multicolumn{5}{c}{\textbf{Code assignment}} \\
\midrule
Llama3-Med42-8B, ICD dict. & 0.229 & 0.231 & 0.23 & 0.13 \\
Llama3-Med42-8B, ICD dict. + \dataset{} & 0.434 & 0.435 & 0.435 & 0.278 \\
Llama3-Med42-8B, ICD dict. + BERGAMOT & 0.403 & 0.405 & 0.404 & 0.253 \\
Qwen2.5-7B-Instruct, ICD dict. & 0.296 & 0.295 & 0.295 & 0.173 \\
Qwen2.5-7B-Instruct, ICD dict. + \dataset{} & 0.456 & 0.449 & 0.452 & 0.292 \\
Qwen2.5-7B-Instruct, ICD dict. + BERGAMOT & 0.305 & 0.303 & 0.304 & 0.179 \\
Phi3\_5\_mini, ICD dict. & 0.394 & 0.39 & 0.392 & 0.244 \\
Phi3\_5\_mini, ICD dict. + \dataset{} & \textbf{0.483} & \textbf{0.477} & \textbf{0.48} & \textbf{0.316} \\
Phi3\_5\_mini, ICD dict. + BERGAMOT & 0.454 & 0.448 & 0.451 & 0.291 \\
Mistral-Nemo, ICD dict. & 0.326 & 0.311 & 0.319 & 0.189 \\
Mistral-Nemo, ICD dict. + \dataset{} & 0.458 & 0.435 & 0.446 & 0.287 \\
Mistral-Nemo, ICD dict. + BERGAMOT & 0.394 & 0.372 & 0.383 & 0.237 \\

\bottomrule
\end{tabular}\caption{\label{tab:llm_tune}ICD coding results for finetuned LLMs on \dataset{}. The best results are highlighted in \textbf{bold}.}
\end{table*}

\section{Implementation Details}\label{appx:impl_details}



\paragraph{Utilized LLMs:} \begin{itemize}
    \item \texttt{Phi-3.5-mini-instruct} \cite{Phi-3.5-mini-instruct}
    \item \texttt{Qwen2.5-7B-Instruct} \cite{Qwen2.5-7B-Instruct}
    \item \texttt{Llama3-Med42-8B} \cite{Med42-8B}
    \item \texttt{Mistral-Nemo-Instruct-2407} \cite{Mistral-Nemo-Instruct-2407}
    \item \texttt{llama3.1:8b-instruct-fp16} \cite{Llama-3.1-8B-Instruct}
\end{itemize}

\paragraph{Diagnosis prediction} Each Longformer was trained for two epochs on separate NVidia A100 GPUs, with the fine-tuning process taking approximately one week per model. We provide hyperparameters for these models training in Tab.~\ref{tab:hp}.

\paragraph{Hyperparameters} A detailed overview, including parameter values and configurations, is provided in Tab.~\ref{tab:hp}.

\begin{table*}[t]
\centering
\begin{tabular}{lrrrr}

\toprule
\multirow{1}{*}{\textbf{Model}} & \multicolumn{1}{c}{\textbf{Precision}}  & \multicolumn{1}{c}{\textbf{Recall}} & \multicolumn{1}{c}{\textbf{F-score}} & \multicolumn{1}{c}{\textbf{Accuracy}}  \\
\midrule 
\multicolumn{5}{c}{\textbf{NER}} \\\midrule
BioBERT, Biosyn, \dataset{} & 0.649 & 0.655 & 0.653 &  0.485 \\
BioBERT, \dataset{} & \textbf{0.721} & \textbf{0.769} & \textbf{0.744} &  \textbf{0.592} \\
BioBERT, NEREL-BIO & 0.588 & 0.675 & 	0.628 &  0.458 \\
BioBERT, NEREL-BIO, \dataset{} & 0.689 & 0.713 & 0.701 &  0.54 \\
BioBERT, RuCCoN & 0.637 & 0.613 & 0.625 &  0.454 \\
BioBERT, RuCCoN + \dataset{} & 0.609 & 0.709 & 0.655 &  0.487 \\
\midrule
\multicolumn{5}{c}{\textbf{NER+Linking}} \\\midrule
BioBERT, Biosyn, \dataset{} & 0.392 & 0.396 & 0.394 &  0.245 \\
BioBERT, \dataset{} & \textbf{0.427} & \textbf{0.455} & \textbf{0.441} &  \textbf{0.283} \\
BioBERT, NEREL-BIO & 0.353 & 0.406 & 0.377 &  0.233 \\
BioBERT, NEREL-BIO, \dataset{} & 0.406 & 0.42 & 0.413 &  0.26 \\
BioBERT, RuCCoN & 0.387 & 0.372 & 0.379 &  0.234 \\
BioBERT, RuCCoN + \dataset{} & 0.351 & 0.409 & 0.378 &  0.233 \\
\midrule
\multicolumn{5}{c}{\textbf{Code assignment}} \\
\midrule
BioBERT, Biosyn, \dataset{} & 0.507 & 0.508 & 0.507 &  0.340 \\
BioBERT, \dataset{} & 0.51 & 0.542 & 	\textbf{0.525} &  \textbf{0.356} \\
BioBERT, NEREL-BIO & 0.466 & 0.531 & 	0.497 &  0.33 \\
BioBERT, NEREL-BIO, \dataset{} & \textbf{0.512} & 0.529 & 	0.52 &  0.352 \\
BioBERT, RuCCoN & 0.508 & 0.485 & 0.496 &  0.33 \\
BioBERT, RuCCoN + \dataset{} & 0.471 & \textbf{0.543} & 0.504 &  0.337 \\
\bottomrule
\end{tabular}\caption{Evaluation results for entity-level tasks for BERT-based IE pipeline on \dataset{} corpus. The best results are highlighted in \textbf{bold}.}\label{tab:ruccod_ie}
\end{table*}

\begin{table*}[t!]
\centering
\begin{tabular}{lrrrr}

\toprule
\multirow{1}{*}{\textbf{Model}} & \multicolumn{1}{c}{\textbf{Precision}}  & \multicolumn{1}{c}{\textbf{Recall}} & \multicolumn{1}{c}{\textbf{F-score}} & \multicolumn{1}{c}{\textbf{Accuracy}}  \\
\midrule 
\multicolumn{5}{c}{\textbf{NER: ICD dict.}} \\\midrule
Llama3.1:8b-instruct & 0.208 & 0.088 & 0.124 & 0.066 \\
Llama3-Med42-8B & 0.202 & 0.084 & 0.118 & 0.063 \\
Phi-3.5-mini-instruct & \textbf{0.211} & \textbf{0.093} & \textbf{0.129} & \textbf{0.069} \\
Mistral-Nemo-Instruct-2407 & 0.198 & 0.072 & 0.105 & 0.055 \\
Qwen2.5-7B-Instruct & 0.206 & 0.087 & 0.122 & 0.065 \\
\midrule 
\multicolumn{5}{c}{\textbf{NER: ICD dict. + \dataset{}}} \\\midrule
Llama3.1:8b-instruct & \textbf{0.581} & \textbf{0.456} & \textbf{0.511} & \textbf{0.343} \\
Llama3-Med42-8B & 0.556 & 0.441 & 0.492 & 0.326 \\
Phi-3.5-mini-instruct & 0.543 & 0.450 & 0.492 & 0.326 \\
Mistral-Nemo-Instruct-2407 & 0.541 & 0.372 & 0.441 & 0.283 \\
Qwen2.5-7B-Instruct & 0.566 & 0.440 & 0.495 & 0.329 \\

\midrule
\multicolumn{5}{c}{\textbf{NER+Linking: ICD dict.}} \\\midrule
Llama3.1:8b-instruct & \textbf{0.071} & 0.067 & \textbf{0.069} & \textbf{0.036} \\
Llama3-Med42-8B & 0.058 & 0.063 & 0.060 & 0.031 \\
Phi-3.5-mini-instruct & 0.062 & \textbf{0.069} & 0.065 & 0.034 \\
Mistral-Nemo-Instruct-2407 & 0.066 & 0.056 & 0.060 & 0.031 \\
Qwen2.5-7B-Instruct & 0.065 & 0.065 & 0.065 & 0.033 \\\midrule
\multicolumn{5}{c}{\textbf{NER+Linking: ICD dict. + \dataset{}}} \\\midrule
Llama3.1:8b-instruct & \textbf{0.272} & \textbf{0.264} & \textbf{0.268} & \textbf{0.155} \\
Llama3-Med42-8B & 0.235 & 0.261 & 0.247 & 0.141 \\
Phi-3.5-mini-instruct & 0.228 & 0.257 & 0.242 & 0.137 \\
Mistral-Nemo-Instruct-2407 & 0.247 & 0.215 & 0.230 & 0.130 \\
Qwen2.5-7B-Instruct & 0.244 & 0.246 & 0.245 & 0.140 \\

\midrule
\multicolumn{5}{c}{\textbf{Code assignment: ICD dict.}} \\
\midrule
Llama3.1:8b-instruct & 0.379 & 0.363 & 0.371 & 0.228 \\
Llama3-Med42-8B & 0.310 & 0.345 & 0.327 & 0.195 \\
Phi-3.5-mini-instruct & 0.260 & 0.294 & 0.276 & 0.160 \\
Mistral-Nemo-Instruct-2407 & \textbf{0.413} & 0.360 & 0.385 & 0.238 \\
Qwen2.5-7B-Instruct & 0.401 & \textbf{0.411} & \textbf{0.406} & \textbf{0.255} \\\midrule
\multicolumn{5}{c}{\textbf{Code assignment: ICD dict. + \dataset{}}} \\\midrule
Llama3.1:8b-instruct & \textbf{0.465} & 0.451 & 0.458 & 0.297 \\
Llama3-Med42-8B & 0.434 & \textbf{0.483} & 0.457 & 0.296 \\
Phi-3.5-mini-instruct & 0.409 & 0.458 & 0.432 & 0.276 \\
Mistral-Nemo-Instruct-2407 & 0.462 & 0.401 & 0.429 & 0.273 \\
Qwen2.5-7B-Instruct & 0.461 & 0.465 & \textbf{0.463} & \textbf{0.301} \\

\bottomrule
\end{tabular}\caption{Evaluation results for NER, Code assignment, and end-to-end entity linking task on RuCCoD for LLM+RAG pipeline.}
\label{rag_res_1}
\end{table*}

\begin{table*}[t!]
\centering
\begin{tabular}{lrrrr}

\toprule
\multirow{1}{*}{\textbf{Model}} & \multicolumn{1}{c}{\textbf{Precision}}  & \multicolumn{1}{c}{\textbf{Recall}} & \multicolumn{1}{c}{\textbf{F-score}} & \multicolumn{1}{c}{\textbf{Accuracy}}  \\
\midrule 
\multicolumn{5}{c}{\textbf{NER:  NEREL-BIO}} \\\midrule
Llama3.1:8b-instruct&0.100&0.042&0.059&0.030\\
Llama3-Med42-8B&0.104&0.043&0.060&0.031\\
Phi-3.5-mini-instruct&0.098&0.043&0.059&0.031\\
Mistral-Nemo-Instruct-2407&\textbf{0.115}&\textbf{0.044}&\textbf{0.063}&\textbf{0.033}\\
Qwen2.5-7B-Instruct&0.099&0.043&0.060&0.031 \\
\midrule 
\multicolumn{5}{c}{\textbf{NER:  RuCCoN}} \\\midrule
Llama3.1:8b-instruct&0.188&0.088&0.120&0.064 \\
Llama3-Med42-8B&0.174&0.079&0.108&0.057 \\
Phi-3.5-mini-instruct&0.172&0.085&0.114&0.060 \\
Mistral-Nemo-Instruct-2407&\textbf{0.197}&0.082&0.116&0.061 \\
Qwen2.5-7B-Instruct&0.185&\textbf{0.091}&\textbf{0.122}&\textbf{0.065} \\

\midrule 
\multicolumn{5}{c}{\textbf{NER+Linking:  NEREL-BIO}} \\\midrule
Llama3.1:8b-instruct&0.023&\textbf{0.020}&0.021&\textbf{0.011} \\
Llama3-Med42-8B&0.018&0.019&0.018&0.009 \\
Phi-3.5-mini-instruct&0.019&\textbf{0.020}&0.019&0.010 \\
Mistral-Nemo-Instruct-2407&\textbf{0.025}&\textbf{0.020}&\textbf{0.022}&\textbf{0.011} \\
Qwen2.5-7B-Instruct&0.021&\textbf{0.020}&0.020&0.010 \\
\midrule 
\multicolumn{5}{c}{\textbf{NER+Linking:  RuCCoN}} \\\midrule
Llama3.1:8b-instruct&0.050&\textbf{0.046}&\textbf{0.048}&\textbf{0.025} \\
Llama3-Med42-8B&0.042&0.044&0.043&0.022 \\
Phi-3.5-mini-instruct&0.038&0.041&0.040&0.020 \\
Mistral-Nemo-Instruct-2407&\textbf{0.053}&0.044&\textbf{0.048}&\textbf{0.025} \\
Qwen2.5-7B-Instruct&0.048&\textbf{0.046}&0.047&0.024 \\

\midrule 
\multicolumn{5}{c}{\textbf{Code assignment:  NEREL-BIO}} \\\midrule
Llama3.1:8b-instruct&0.059&0.053&0.056&\textbf{0.029} \\ 
Llama3-Med42-8B&0.045&0.047&0.046&0.024 \\
Phi-3.5-mini-instruct&0.046&0.049&0.047&0.024 \\
Mistral-Nemo-Instruct-2407&\textbf{0.062}&0.051&0.056&\textbf{0.029} \\
Qwen2.5-7B-Instruct&0.058&\textbf{0.056}&\textbf{0.057}&\textbf{0.029} \\
\midrule 
\multicolumn{5}{c}{\textbf{Code assignment:  RuCCoN}} \\\midrule
Llama3.1:8b-instruct&\textbf{0.164}&\textbf{0.150}&\textbf{0.157}&\textbf{0.085} \\
Llama3-Med42-8B&0.125&0.131&0.128&0.068 \\
Phi-3.5-mini-instruct&0.125&0.134&0.129&0.069 \\
Mistral-Nemo-Instruct-2407&0.156&0.129&0.141&0.076 \\
Qwen2.5-7B-Instruct&0.156&0.152&0.154&0.084 \\

\bottomrule
\end{tabular}\caption{Evaluation results for NER, Code assignment, and end-to-end entity linking task on RuCCoD for LLM+RAG pipeline using NEREL-BIO and RuCCoN for vectorstore.}
\label{rag_res_2}
\end{table*}

\begin{table*}[t!]
\centering
\begin{tabular}{lrrrr}

\toprule
\multirow{1}{*}{\textbf{Model}} & \multicolumn{1}{c}{\textbf{Precision}}  & \multicolumn{1}{c}{\textbf{Recall}} & \multicolumn{1}{c}{\textbf{F-score}} & \multicolumn{1}{c}{\textbf{Accuracy}}  \\
\midrule 
\multicolumn{5}{c}{\textbf{NER: ICD dict.}} \\\midrule
Llama3.1:8b-instruct&0.208&0.088&0.124&0.066 \\
Llama3-Med42-8B&0.202&0.084&0.118&0.063 \\
Phi-3.5-mini-instruct&\textbf{0.211}&\textbf{0.093}&\textbf{0.129}&\textbf{0.069} \\
Mistral-Nemo-Instruct-2407&0.198&0.072&0.105&0.055 \\
Qwen2.5-7B-Instruct&0.206&0.087&0.122&0.065 \\
\midrule 
\multicolumn{5}{c}{\textbf{NER: ICD dict. + \dataset{}}} \\\midrule
Llama3.1:8b-instruct&\textbf{0.581}&\textbf{0.456}&\textbf{0.511}&\textbf{0.343} \\
Llama3-Med42-8B&0.556&0.441&0.492&0.326 \\
Phi-3.5-mini-instruct&0.543&0.450&0.492&0.326 \\
Mistral-Nemo-Instruct-2407&0.541&0.372&0.441&0.283 \\
Qwen2.5-7B-Instruct&0.566&0.440&0.495&0.329 \\

\midrule
\multicolumn{5}{c}{\textbf{NER+Linking: ICD dict.}} \\\midrule
Llama3.1:8b-instruct&\textbf{0.095}&0.088&\textbf{0.091}&\textbf{0.048} \\
Llama3-Med42-8B&0.077&0.083&0.080&0.042 \\
Phi-3.5-mini-instruct&0.083&\textbf{0.092}&0.087&0.046 \\
Mistral-Nemo-Instruct-2407&0.083&0.070&0.076&0.040 \\
Qwen2.5-7B-Instruct&0.087&0.086&0.087&0.045 \\
\midrule
\multicolumn{5}{c}{\textbf{NER+Linking: ICD dict. + \dataset{}}} \\\midrule
Llama3.1:8b-instruct&\textbf{0.378}&\textbf{0.362}&\textbf{0.369}&\textbf{0.227} \\
Llama3-Med42-8B&0.324&0.354&0.338&0.203 \\
Phi-3.5-mini-instruct&0.323&0.357&0.339&0.204 \\
Mistral-Nemo-Instruct-2407&0.342&0.295&0.317&0.188 \\
Qwen2.5-7B-Instruct&0.343&0.340&0.342&0.206 \\

\midrule
\multicolumn{5}{c}{\textbf{Code assignment: ICD dict.}} \\
\midrule
Llama3.1:8b-instruct&0.575&0.561&0.568&0.396 \\
Llama3-Med42-8B&0.523&0.594&0.556&0.385 \\
Phi-3.5-mini-instruct&0.437&0.510&0.471&0.308 \\
Mistral-Nemo-Instruct-2407&\textbf{0.598}&0.533&0.564&0.392 \\
Qwen2.5-7B-Instruct&0.595&\textbf{0.618}&\textbf{0.607}&\textbf{0.435} \\
\midrule
\multicolumn{5}{c}{\textbf{Code assignment: ICD dict. + \dataset{}}} \\\midrule
Llama3.1:8b-instruct&\textbf{0.701}&0.684&0.692&0.529 \\
Llama3-Med42-8B&0.644&\textbf{0.720}&0.680&0.515 \\
Phi-3.5-mini-instruct&0.627&0.703&0.663&0.496 \\
Mistral-Nemo-Instruct-2407&0.691&0.605&0.645&0.476 \\
Qwen2.5-7B-Instruct&0.700&0.704&\textbf{0.702}&\textbf{0.541} \\

\bottomrule
\end{tabular}\caption{Relaxed evaluation results for NER, Code assignment, and end-to-end entity linking task on RuCCoD for LLM+RAG pipeline.}
\label{rag_res_3}
\end{table*}

\begin{table*}[t!]
\centering
\begin{tabular}{lrrrr}

\toprule
\multirow{1}{*}{\textbf{Model}} & \multicolumn{1}{c}{\textbf{Precision}}  & \multicolumn{1}{c}{\textbf{Recall}} & \multicolumn{1}{c}{\textbf{F-score}} & \multicolumn{1}{c}{\textbf{Accuracy}}  \\


\midrule 
\multicolumn{5}{c}{\textbf{NER:  NEREL-BIO}} \\\midrule
Llama3.1:8b-instruct-fp16&0.100&0.042&0.059&0.030 \\
Llama3-Med42-8B&0.104&0.043&0.060&0.031 \\
Phi-3.5-mini-instruct&0.098&0.043&0.059&0.031 \\
Mistral-Nemo-Instruct-2407&\textbf{0.115}&\textbf{0.044}&\textbf{0.063}&\textbf{0.033} \\
Qwen2.5-7B-Instruct&0.099&0.043&0.060&0.031 \\

\midrule 
\multicolumn{5}{c}{\textbf{NER:  RuCCoN}} \\\midrule
Llama3.1:8b-instruct-fp16&0.188&0.088&0.120&0.064 \\
Llama3-Med42-8B&0.174&0.079&0.108&0.057 \\
Phi-3.5-mini-instruct&0.172&0.085&0.114&0.060 \\
Mistral-Nemo-Instruct-2407&\textbf{0.197}&0.082&0.116&0.061 \\
Qwen2.5-7B-Instruct&0.185&\textbf{0.091}&\textbf{0.122}&\textbf{0.065} \\

\midrule 
\multicolumn{5}{c}{\textbf{NER+Linking:  NEREL-BIO}} \\\midrule
Llama3.1:8b-instruct&\textbf{0.033}&\textbf{0.029}&\textbf{0.031}&\textbf{0.016} \\
Llama3-Med42-8B&0.024&0.025&0.025&0.013 \\
Phi-3.5-mini-instruct&0.026&0.028&0.027&0.014 \\
Mistral-Nemo-Instruct-2407&\textbf{0.033}&0.027&0.030&0.015 \\
Qwen2.5-7B-Instruct&0.030&\textbf{0.029}&0.030&0.015 \\
\midrule 
\multicolumn{5}{c}{\textbf{NER+Linking:  RuCCoN}} \\\midrule
Llama3.1:8b-instruct&\textbf{0.076}&0.069&\textbf{0.072}&\textbf{0.038} \\
Llama3-Med42-8B&0.061&0.063&0.062&0.032 \\
Phi-3.5-mini-instruct&0.060&0.064&0.062&0.032 \\
Mistral-Nemo-Instruct-2407&\textbf{0.076}&0.062&0.068&0.035 \\
Qwen2.5-7B-Instruct&0.073&\textbf{0.070}&\textbf{0.072}&0.037 \\

\midrule 
\multicolumn{5}{c}{\textbf{Code assignment:  NEREL-BIO}} \\\midrule
Llama3.1:8b-instruct&0.114&0.107&0.110&0.058 \\
Llama3-Med42-8B&0.088&0.096&0.092&0.048 \\
Phi-3.5-mini-instruct&0.098&0.110&0.104&0.055 \\
Mistral-Nemo-Instruct-2407&0.121&0.105&0.112&0.059 \\
Qwen2.5-7B-Instruct&\textbf{0.125}&\textbf{0.126}&\textbf{0.125}&\textbf{0.067} \\
\midrule 
\multicolumn{5}{c}{\textbf{Code assignment:  RuCCoN}} \\\midrule
Llama3.1:8b-instruct&\textbf{0.295}&0.282&0.288&0.168 \\
Llama3-Med42-8B&0.254&0.275&0.264&0.152 \\
Phi-3.5-mini-instruct&0.248&0.273&0.260&0.149 \\
Mistral-Nemo-Instruct-2407&0.284&0.244&0.263&0.151 \\
Qwen2.5-7B-Instruct&0.292&\textbf{0.294}&\textbf{0.293}&\textbf{0.172} \\

\bottomrule
\end{tabular}\caption{Relaxed evaluation results for NER, Code assignment, and end-to-end entity linking task on RuCCoD for LLM+RAG pipeline using NEREL-BIO and RuCCoN for vectorstore.}
\label{rag_res_4}
\end{table*}


\section{Prompts}\label{sec:prompt}

The original prompts were in Russian. Below are their translations to English.

\begin{tcolorbox}[
  enhanced,
  breakable,
  title=NER prompt,
  width=\linewidth,        
  listing only,
  listing options={
    basicstyle=\footnotesize\ttfamily,
    breaklines=true,
    breakatwhitespace=true,
    numbers=none,
    xleftmargin=0pt,
    showstringspaces=false
  }
]
You will be provided with a text containing diagnoses. Extract the diagnoses from this text. Do not alter the spelling of the diagnoses in the text. Respond only in the format of a list: ['diagnosis1', 'diagnosis2', ...]
Text: \{text\}
\label{ner-prompt}
\end{tcolorbox}

\begin{tcolorbox}[
  enhanced,
  breakable,
  title=Diagnosis selection prompt,
  width=\linewidth,
  listing only,
  listing options={
    basicstyle=\footnotesize\ttfamily,
    breaklines=true,
    breakatwhitespace=true,
    numbers=none,
    xleftmargin=0pt,
    showstringspaces=false
  }
]
You will be given a reference diagnosis and a list of diagnoses from a database.
Your task is to determine which diagnosis from the database best matches the reference diagnosis.
Try to select the diagnosis accurately, paying attention to details. Choose the diagnosis with the highest match in terms of words and meaning.
You can only choose from the diagnoses in the list.
Pay more attention to the diagnoses at the beginning of the list, as they are more likely to be a better match.
It's better to choose a shorter diagnosis than one that includes information not present in the reference diagnosis.
In your response, write only the diagnosis number and nothing else.
Reference diagnosis: \{diagnosis\}
List of diagnoses from a database: \{list\}
\end{tcolorbox}

\end{document}